\documentclass[12pt,a4paper,DVI=10]{article}
\usepackage[utf8]{inputenc}
\usepackage{lmodern}
\usepackage{comment}
\usepackage{lscape}
\usepackage{amsmath}	
\usepackage{amssymb}
\usepackage{amsthm}
\usepackage{cases}
\usepackage{graphicx}
\usepackage{caption}
\usepackage[table]{xcolor}

\usepackage{multirow}
\usepackage{amsfonts}
\usepackage{float} 
\usepackage{diagbox}
\usepackage{makecell}
\usepackage{fullpage}
\usepackage{url}
\usepackage{rotating}
\usepackage{eurosym}
\usepackage{wrapfig}
\usepackage[final]{pdfpages} 
\usepackage{epstopdf} 
\usepackage{subfig} 
\usepackage[a4paper]{geometry}
\usepackage[nottoc]{tocbibind} 
\usepackage{placeins}
\usepackage{float}

\usepackage{listings}
\usepackage{color, colortbl}

\usepackage{hyperref}
\usepackage{xcolor}
\usepackage{titlesec}
\usepackage[tableposition=top]{caption}

\usepackage{graphicx}
\usepackage{lscape}
\usepackage{caption}
\usepackage{comment}

\usepackage{makecell}
\usepackage{multirow}
\usepackage{amsfonts}
\usepackage{float} 
\usepackage{diagbox}
\usepackage{url}
\usepackage{rotating}
\usepackage{eurosym}
\usepackage{wrapfig}
\usepackage{epstopdf}
\usepackage[nottoc]{tocbibind}
\usepackage{placeins}
\usepackage{listings}
\usepackage{color, colortbl}
\usepackage{hyperref}
\usepackage{xcolor}
\usepackage{sidecap}
\usepackage{booktabs}
\usepackage{stmaryrd}
\usepackage{makecell}
\usepackage[utf8]{inputenc}
\usepackage{lmodern}
\usepackage[T1]{fontenc}
\usepackage{lscape}
\usepackage{amsmath}	
\usepackage{amssymb}
\usepackage{stmaryrd}
\usepackage{scalerel}
\usepackage{subfig}
\usepackage{algpseudocode}
\usepackage{algorithm}
\usepackage{setspace}
\usepackage{tikz-cd}
\DeclareGraphicsExtensions{.eps}

\usepackage[printonlyused]{acronym}

\newcommand{\rev}{\textcolor{black}}

\newcommand{\bx}{\textbf{x}}

\newcommand{\by}{\textbf{y}}
\newcommand{\bB}{\textbf{B}}
\newcommand{\bH}{\textbf{H}}
\newcommand{\bR}{\textbf{R}}
\newcommand{\bA}{\textbf{A}}

\newcommand{\bD}{\textbf{D}}

\newcommand{\bC}{\textbf{C}}
\newcommand{\bL}{\textbf{L}}
\newcommand{\bV}{\textbf{V}}

\newcommand{\bW}{\textbf{W}}

\newcommand{\bX}{\textbf{X}}

\newcommand{\bI}{\textbf{I}}

\newcommand{\bb}{\textbf{b}}
\newcommand{\bu}{\textbf{u}}

\newcommand{\balpha}{\boldsymbol{\alpha}}
\newcommand{\argmin}{\operatornamewithlimits{argmin}}
\newcommand{\sibo}{\textcolor{blue}}

\makeatletter
\AtBeginDocument{%
  \renewcommand*{\AC@hyperlink}[2]{%
    \begingroup
      \hypersetup{hidelinks}%
      \hyperlink{#1}{#2}%
    \endgroup
  }%
}
\makeatother
\begin{document}

\title{Generalised Latent Assimilation in Heterogeneous Reduced Spaces with Machine Learning Surrogate Models
}

\author{Sibo Cheng$^{1}$, Jianhua Chen$^{2,5}$, Charitos Anastasiou$^{3}$,  Panagiota Angeli$^{3}$, \\Omar K. Matar$^{2}$, Yi-Ke Guo$^{1}$, Christopher C. Pain$^{4}$, Rossella Arcucci$^{1,4}$ \\
\\
        \small $^{1}$ Data Science Instituite, Department of Computing, Imperial College London, UK \\
        \small $^{2}$ Department of Chemical Engineering, Imperial College London, UK\\
        \small $^{3}$ Department of Chemical Engineering, University College London, UK\\
        \small $^{4}$ Department of Earth Science \& Engineering, Imperial College London, UK\\
        \small $^{5}$ State Key Laboratory of Multiphase Complex Systems, Institute of Process Engineering,\\ \small Chinese Academy of Sciences, China\\
}


\maketitle

\abstract{Reduced-order modelling and low-dimensional surrogate models generated using machine learning algorithms have been widely applied in high-dimensional dynamical systems to improve the \rev{algorithmic efficiency}. In this paper, we develop a system which combines reduced-order surrogate models with a novel data assimilation (DA) technique used to incorporate real-time observations from different physical spaces. We make use of local smooth surrogate functions which link the space of encoded system variables and the one of current observations to perform variational DA with a low computational cost. The new system, named Generalised Latent Assimilation can benefit both the efficiency provided by the \rev{reduced-order modelling} and the accuracy of data assimilation.
A theoretical analysis of the difference between surrogate and original assimilation cost function is also provided in this paper where an upper bound, depending on the size of the local training set, is given. 
 The new approach is tested \rev{on} a high-dimensional \ac{CFD} application of a two-phase liquid flow with non-linear observation operators that current Latent Assimilation methods can not handle. 
 Numerical results demonstrate that the proposed assimilation approach can significantly improve the reconstruction and prediction accuracy of the deep learning surrogate model which is nearly 1000 times faster than the CFD simulation.}




\section{Introduction}
Spatial field prediction and reconstruction are crucial in the control of high-dimensional physical systems for applications in 
\ac{CFD}, geoscience or medical science.    
Running physics-informed simulations is often computationally expensive, especially for high resolution and multivariate systems. 
Over the past years, numerous studies have been devoted to speed up the simulation/prediction of dynamical systems by constructing surrogate models via \ac{ROM} and \ac{ML} techniques~\cite{nakamura2021convolutional, mohan2018deep, casas2020,fu2021data}. More precisely, the simulation/experimental data are first compressed to a low-dimensional latent space through an \ac{AE}. A \ac{RNN} is then used to train a \rev{reduced-order} surrogate model for predicting the dynamics in the latent space using compressed data. Once the \ac{ML} surrogate model is computed, monitoring the model prediction with limited sensor information constitutes another major challenge. 
Making use of a weighted combination of simulation (also known as `background') and observation data~\cite{Carrassi2017}, \ac{DA} methods are widely used in engineering applications for field prediction or parameter identification~\cite{Carrassi2017,Gong2020}. 

To incorporate real-time observations for correcting the prediction of the surrogate model, the idea of \ac{LA} was introduced~\cite{amendola2020,peyron2021latent,silva2021data} where \ac{DA} is performed directly in the reduced-order latent space.  It has been shown in~\cite{amendola2020} that \ac{LA} has a significant advantage in terms of computational efficiency compared to classical full-space \ac{DA} methods. However, current approaches of \ac{LA} require the compression of the observation data into the same latent space of the state variables, which is cumbersome for some applications where the states and the observations are either compressed using different \acp{AE} or different physical quantities. The latter is  common practice in geoscience and \ac{CFD} applications. For example, the observation of wind speed/direction can be used to improve the quality of the initial conditions of weather forecasts~\cite{fowler2018interaction} and precipitation data can be used to correct the river flow prediction in hydrology~\cite{cheng2020b,cheng2021observation}.

\rev{The \ac{DA} is performed through a transformation operator (usually denoted by $\mathcal{H}$) which links the state variables to real-time observations. In real applications, $\mathcal{H}$ is often highly non-linear~\cite{Nichols2010}. In the case of \ac{LA}, since the assimilation is carried out in the latent space, the $\mathcal{H}$ function also includes several encoder, decoder functions, leading to extra difficulties in solving the assimilation problem.}
Furthermore, if the state vector and the observation vector are not in the same physical space, the latent spaces where the data are reduced might be different too. In this case, the operator of the data assimilation inverse problem includes the two \ac{ML}-based functions used to compress the data (state vector and observations) in two different latent spaces. 
Also, \ac{ML} functions often involve many parameters and are difficult to train in real-time. This means that performing variational \ac{LA}, when the background simulation and the observation vector are not in the same physical space, is cumbersome.

The idea of applying \ac{ML} algorithms, namely recurrent neural networks in a low-dimensional latent space for learning complex dynamical systems has been recently adapted in a wide range of applications including CFD~\cite{san2019,mohan2018deep}, hydrology~\cite{cheng2021observation}, nuclear science~\cite{gong2022data} and air pollution quantification~\cite{casas2020}. Both \rev{\ac{POD}-type} (e.g.,~\cite{mohan2018deep,Arcucci2018,casas2020,cheng2021observation}) and \acp{NN}-based autoencoding methods~\cite{san2019,nakamura2021convolutional} have been used to construct the reduced-order latent spaces. The work of~\cite{casas2020} is extended in~\cite{quilodran2021adversarial} which relies on an Adversarial \ac{RNN} when the training dataset is insufficient. In terms of compression accuracy, much effort has been devoted to compare the performance of different auto-encoding approaches. The study of~\cite{murata2020nonlinear} shows a significant advantage of \acp{NN}-based methods compared to classical POD-type approaches when dealing with highly non-linear \ac{CFD} applications. 
A novel \ac{ROM} method, combining \ac{POD} and \acp{NN} \ac{AE} has been introduced in the very recent work of~\cite{phillips2021autoencoder}. The authors have demonstrated that one of the advantages of this approach, for projection-based ROMs,
is that it does not matter whether the high-fidelity solution is on a structured or unstructured mesh. \rev{Other approaches applying convolutional autoencoders} to data on unstructured meshes include space-filling curves~\cite{heaney2020applying}, spatially varying kernels~\cite{zhou2020fully} or graph-based networks~\cite{xu2021ucnn}.

Performing DA in the latent space in order to monitor surrogate models with real-time observations  has led to an increase in research interest recently. The approaches used in the work of~\cite{casas2020,arcucci2021deep} consist of learning assimilated results directly via a \ac{RNN} to reduce forecasting errors. With a similar idea, \cite{Brajard2020} proposes an iterative process of \ac{DL} and \ac{DA}, i.e., a \ac{NN} is  retrained after each DA step (based on \ac{NN} predictions and real observations) until convergence has been achieved. Collectively, the methods in \cite{casas2020,arcucci2021deep,Brajard2020} aim to enhance the system prediction by including assimilated dynamics in the training data. However, the requirement to retrain the \ac{NN} when new observation data become available leads to considerable computational cost for online application of these methods.

In order to  incorporate unseen real-time observation data efficiently, the recent works of~\cite{amendola2020, peyron2021latent,LIU202246} introduce the concept of \ac{LA} where an \ac{AE} network is used to compress the state variables and pre-processed observation data. The \ac{DA} updating is performed in the reduced-order latent space subsequently. Similarly, in~\cite{silva2021data}, a Generative Adversarial Network (GAN) was trained to produce time series data of \ac{POD} coefficients, and this algorithm was extended to assimilate data by modifying the loss function and using the back-propagation algorithm of the GAN. Again, this produces an efficient method as no additional simulations of the high-fidelity model are required during the data assimilation process.
Also, \cite{becker2019} proposes the use of a recurrent Kalman network in the latent space to make locally linear predictions. However, as mentioned in the Introduction, an important bottleneck of the current \ac{LA} techniques is that the state and observation variables often can not be encoded into the same latent space for complex physical systems. Performing online \ac{LA} thus requires a smooth, explainable and efficient-to-train local surrogate transformation function, leading to our idea of implementing polynomial regression.

Local polynomial regression has been widely used for the prediction and calibration of chaotic systems by providing smooth and easily interpretable surrogate functions. The work of~\cite{su2010prediction} uses multivariate local polynomial fitting (M-MLP) which takes previous time steps in a multivariate dynamical systems as input and forecasts the evolution of the state variables. It is  demonstrated numerically that the M-MLP outperforms a standard \ac{NN} in the Lorenz twin experiment.  Recently this work has been developed by the same authors to a local polynomial autoregressive model \cite{su2015local} which shows a good performance in one-step prediction. A detailed numerical comparison between \ac{PR} and \ac{NN} has also been given in \cite{choon2008functional, hwang2021deep}. Their results show that \ac{PR} with a polynomial degree lower than five, can achieve similar results to \acp{NN} when fitting a variety of multivariate real functions. Using a similar idea, \cite{regonda2005local} applies the local polynomial regression to provide not only the single mean forecast but an ensemble of future time steps, which provides better forecasts with noisy data as proved in their paper with geological applications. 

Polynomial regression, or more generally, interpretable surrogate models such as Lasso or a Decision Tree (DT), have been widely used to approximate sophisticated deep learning algorithms to improve interpretability \cite{molnar2020interpretable}. For example, \cite{ribeiro2016should} developed the model of Local Interpretable Model-agnostic Explanations (LIME) for improving the interpretability of \ac{ML} classifiers. More precisely, they make use of a linear regression model to approximate a \acp{NN} classifier where the loss function is defined as a fidelity-interpretability tradeoff. The training set of the linear surrogate model is generated via samplings for local exploration of each \ac{ML} input. It is pointed out by both \cite{molnar2020interpretable} and \cite{ribeiro2016should} that both the distribution and the range of local samplings are crucial to the robustness of the local surrogate model. A small range may lead to  overfitting while the efficiency and the local fidelity can decrease when the sampling range is too large. 

A graph-based sampling strategy is proposed in the recent work of \cite{shi2020modified} to improve the performance of LIME. The principle of LIME can be easily extended by using a polynomial regression since our prime concern is not the interpretability but the smoothness of the local surrogate model.
On the other hand, some effort has been given to replace the computational expensive \ac{ML} models by polynomial functions which are much more efficient to evaluate. The use of a data-driven polynomial chaos expansion (PCE) has been proposed recently by \cite{torre2019data} to perform \ac{ML} regression tasks with a similar performance compared to DL and Support vector machine. Furthermore, PCE is able to deliver a probability density function instead of a single mean prediction for the model output. A similar idea can be found in \cite{shahzadi2021deep} where the authors compare PCE- and \acp{NN}-based surrogate models for sensitivity analysis in a real-world geophysical problem. The study of \cite{emschwiller2020neural} aims to reduce the over-parametrization of neural networks by using polynomial functions to fit a trained \ac{NN} of the same inputs. Their study includes sophisticated \acp{NN} structures such as \ac{2D} \ac{CNN}, in the global space. 
Despite the fact that the classification accuracy of the surrogate polynomial regression is slightly lower than the state-of-the-art DL approaches, the former exhibits a significantly higher noise robustness on real datasets. In addition, the theoretical study in \cite{emschwiller2020neural} provides an upper bound of the \ac{PR} \rev{learning error with respect to the number of samplings}.
Another important advantage of \ac{PR} compared to other \ac{ML} models, namely deep learning approaches, is the good performance for small training sets thanks to the small number of tuning parameters required \cite{torre2019data}. Moreover, unlike DL methods, polynomial regression requires much less fine tuning of hyper-parameters which makes it more appropriate for online training tasks. 

\begin{figure}[h!]
\centering
\includegraphics[width = 4.7in]{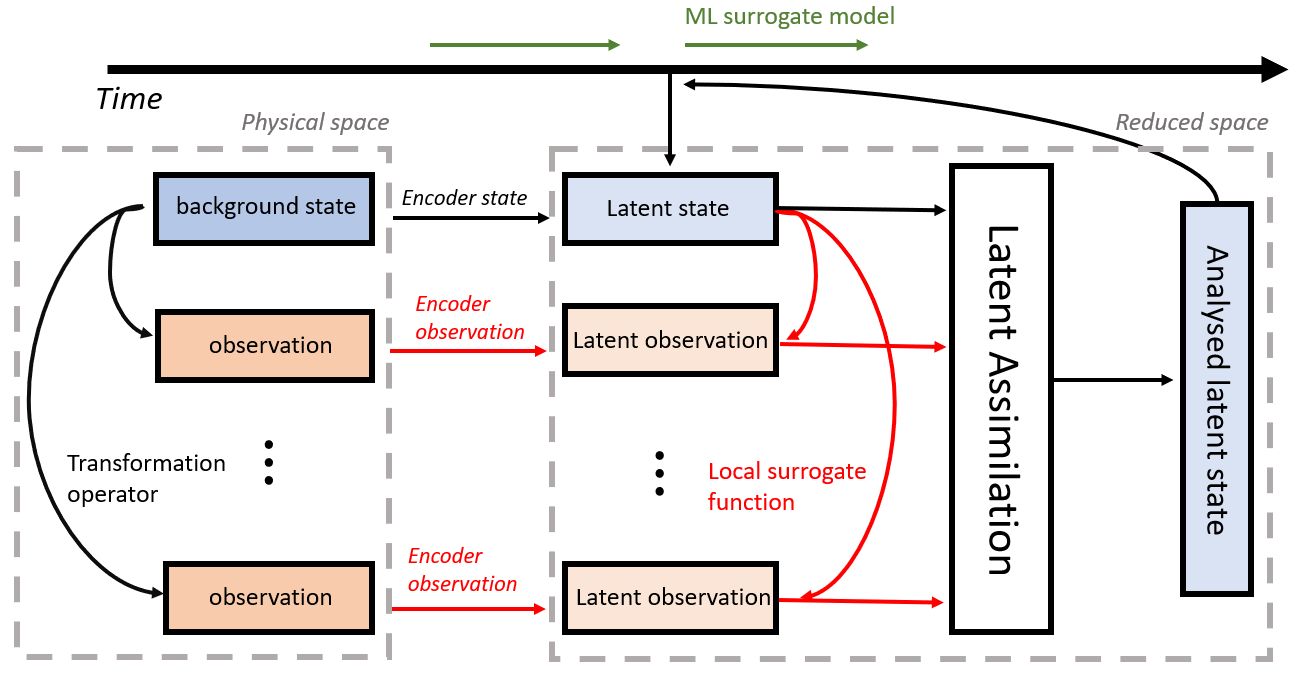} 
   \caption{Flowchart of the Generalised Latent Assimilation with machine learning surrogate models.}
   \label{fig:flowchart_PR}
\end{figure}

In this study, we develop a novel \ac{LA} algorithm scheme which generalises the current \ac{LA} framework~\cite{amendola2020} to heterogeneous latent spaces and non-linear transformation operators while keeping the important advantage of \ac{LA} in terms of low computational cost. We use local surrogate functions to approximate the transformation operator from the latent space of the state vector to the observation one. This approach can incorporate observation data from different sources in one assimilation window as shown in Figure~\ref{fig:flowchart_PR}. The latent transformation operator, which combines different encoder/decoder networks, and the state-observation transformation mapping, $\mathcal{H}$ in the full physical space, is then used to
solve the \ac{LA} inverse problem.
A crucial requirement is ensuring both the approximation accuracy (for unseen data) and the \rev{smoothness and interpretability} of the surrogate function. For these reasons, we used local \ac{PR} which is sufficiently accurate and infinitely differentiable~\cite{ostertagova2012modelling}. We provide both a theoretical and numerical analysis (based on a high-dimensional \ac{CFD} application) of the proposed method. The surrogate models we build are based \rev{on \ac{AE}} and \ac{LSTM} technologies which have been shown to provide stable and accurate solutions for \acp{ROM}~\cite{quilodran2021adversarial}. \\

In summary, we make the following contributions in this study:
\begin{itemize}
    \item We propose a novel  Generalised Latent Assimilation algorithm. Making use of a local \ac{PR} to open the blackbox of \ac{DL} functions addresses one of the major bottlenecks of current \ac{LA} approaches for combining information sources (namely state vector and observations) issued from different latent spaces. The main differences of the proposed novel Generalised \ac{LA} compared to the existing \ac{LA} approaches are underlined in red in Figure~\ref{fig:flowchart_PR}.
    
    \item We provide a theoretical error upper-bound for the expectation of the cost function in \ac{LA} when using the local surrogate polynomial function instead of the original \ac{DL} function. This upper-bound, depending on the polynomial degree and the input dimension, is obtained based on the general results of learning \acp{NN} functions via \ac{PR} \cite{emschwiller2020neural}. 

    \item The new approach proposed in this work is general and it can be easily applied/extended to other dynamical systems.
\end{itemize}

The rest of this paper is organised as follows. In Section~\ref{sec:ROM}, several dimension reduction methods, including \ac{POD}, ML-based \ac{AE} and \ac{POD} \ac{AE} are introduced. We then address the \ac{RNN} latent surrogate model in Section~\ref{sec:LSTM}. The novel Generalised \ac{LA} approach with a theoretical analysis is described in Section~\ref{sec:latent assimilation} after the introduction of classical variational \ac{DA}. The \ac{CFD} application, as a test case in this paper, is briefly explained in Section~\ref{sec:CFD}. The numerical results of this study are split into two parts: Section~\ref{sec:results1} for latent surrogate modelling (including \ac{ROM} reconstruction and \ac{LSTM} prediction), and Section~\ref{sec:results2} for Generalised \ac{LA} with heterogeneous latent spaces. Finally, concluding remarks are provided in Section~\ref{sec:conclusions}.

\section{Methodology: \ac{ROM} and RNN}
\subsection{Reduced-order-modelling}
\label{sec:ROM}
Different \ac{ROM} approaches are introduced in this section with the objective to build an efficient rank reduction model with a low dimensional latent space and high accuracy of reconstruction. Their performance is later compared in the oil-water flow application in Section~\ref{sec:numerical_AE}.

\subsubsection{Proper orthogonal decomposition}
\label{sec: POD}
The principle of proper orthogonal decomposition was introduced in the work of \cite{lumley1967structure}. 
 In general, a set of $n_\textrm{state}$ state snapshots, issued from one or several simulated or observed dynamics, is represented by a matrix $\bX \in \mathbb{R}^{[\textrm{dim}(\bx) \times n_\textrm{state}]}$ where \rev{each column of $\bX$} represents an individual state vector \rev{at a given time instant} (also known as snapshots), i.e.
\begin{align}
    \bX [:,i] = \bx_{t=t_i}, \quad \forall i \in \rev{\{0,1,..., n_\textrm{state}-1\}}.
\end{align}
Thus the ensemble $\bX$ describes the evolution of the state vectors. Its empirical covariance $\bC_{\bx}$ can be written and decomposed as 

\begin{align}
    \bC_{\bx} = \frac{1}{n_\textrm{state}-1} \bX \bX^T = {\bL}_{\bX} {\bD}_{\bX} {{\bL}_{\bX}}^T
\end{align}
where the columns of ${\bL}_{\bX}$ are the principal components of $\bX$ and ${\bD}_{\bX}$ \rev{is a diagonal matrix collecting} the associated eigenvalues $\{ \lambda_{\bX,i}, i=0,...,n_\textrm{state}-1\}$ in a decreasing order, i.e.,
\begin{align}
  {\bD}_{\bX} =
  \begin{bmatrix}
    \lambda_{\bX,0} & & \\
    & \ddots & \\
    & & \lambda_{\bX,n_\textrm{state}-1}
  \end{bmatrix}.
\end{align}
For a truncation parameter \rev{$q \leq n_\textrm{state}$}, one can construct a projection operator ${\bL}_{\bX,q}$ with minimum loss of information by keeping the \rev{first $q$ columns of} ${\bL}_{\bX}$. This projection operator can also be obtained by a \ac{SVD} \cite{stewart1993early} which does not require computing the full covariance matrix $\bC_{\bx}$. More precisely,
\begin{align}
     \bX = {\bL}_{\bX,q} \boldsymbol{\Sigma} {\bV}_{\bX,q}  \label{eq:SVD_X}
\end{align}
where ${\bL}_{\bX,q}$ and ${\bV}_{\bX,q}$ are \rev{by definition with orthonormal columns}. , i.e.,
\begin{align}
    {{\bL}_{\bX,q}}^T {\bL}_{\bX,q} = {{\bV}_{\bX,q}}^T {{\bV}_{\bX,q}} = \bI \quad \textrm{and} \quad \boldsymbol{\Sigma}\boldsymbol{\Sigma}^T = {\bD}_{q,X},
\end{align}
 \rev{where ${\bD}_{q,X}$ is a diagonal matrix containing the first $q$ eigenvalues of ${\bD}_{X}$}. For a single state vector $\bx$, the compressed latent vector $\tilde{\bx}$ can be written as
\begin{align}
    \tilde{\bx} =  {{\bL}_{\bX,q}}^T \bx, \label{eq: reconstruction}
\end{align}
which is a reduced rank approximation to the full state vector $\bx$. \rev{The POD reconstruction then reads,}
\begin{align}
    \bx^r_\textrm{POD} = {{\bL}_{\bX,q}} \tilde{\bx} = {{\bL}_{\bX,q}} {{\bL}_{\bX,q}}^T \bx.
\end{align}
The compression rate $\rho_{\bx}$ and the compression accuracy $\gamma_{\bx}$ are defined respectively as:
\begin{align}
    \gamma_{\bx} = \sum_{i=0}^{q-1} \lambda^2_{\bX,i} \Big/ \sum_{i=0}^{n_\textrm{state}-1} \lambda^2_{\bX,i}  \quad \textrm{and} \quad \rho_{\bx} = q \big/ n_\textrm{state}. \label{eq:POD rate}
\end{align}

\subsubsection{Convolutional auto-encoder}
\label{sec:CAE}
An auto-encoder is a special type of artificial \acp{NN} used to perform data compression via an unsupervised learning of the identity map. The network structure of an \ac{AE} can be split into two parts: an encoder which maps the input vector to the latent space, and a decoder which connects the latent space and the output.  
More precisely, the encoder $\mathcal{E}_x$ first encodes the inputs $\bx$ to latent vector $\tilde{\bx} = \mathcal{E}_{\bx} (\bx)$, which is often of a much lower dimension (i.e., $\textrm{dim}(\tilde{\bx}) \ll \textrm{dim}(\bx) $). A decoder $\mathcal{D}_{\bx} $ is then added to approximate the 
input vector $\bx$ by computing a reconstructed vector $\bx^r_\textrm{AE} = \mathcal{D}_{\bx} \big( \mathcal{E}_{\bx} (\bx) \big)$.
 The encoder and the decoder are then trained jointly with, for instance, the \ac{MSE} \rev{as the loss function}
\begin{align}
J\big( \boldsymbol{\theta}_{\mathcal{E}}, \boldsymbol{\theta}_{\mathcal{D}}\big) = \frac{1}{N_\textrm{train}^{\textrm{AE}}} \sum_{j=1}^{N_\textrm{train}^{\textrm{AE}}} \vert \vert\bx_j - \bx^\textrm{r}_{\textrm{AE},j}\vert \vert^2
\end{align}
where $ \boldsymbol{\theta}_{\mathcal{E}}, \boldsymbol{\theta}_{\mathcal{D}}$ denote the parameters in the encoder and the decoder respectively, and $N_\textrm{train}^{\textrm{AE}}$ represents the size of the \ac{AE} training dataset. 

Neural networks with additional layers or more sophisticated structures (e.g.,CNN or RNN) can better recognise underlying spatial or temporal patterns, resulting in a more effective representation of complex data. Since we aim to obtain a static encoding (i.e., a single latent vector will not contain temporal information) at this stage, we make use of a CNN to build our first AE. In general, a convolutional layer makes use of a local filter to compute the values in the next layer. By shifting the input tensor \rev{by a convolutional window} of fixed size, we obtain the output of a convolutional layer~\cite{rawat2017deep}.
 Compared to standard \ac{AE} with dense layers, the advantage of \ac{CAE} is mainly two-folds: the reduction of the number of parameters in the \ac{AE} and the capability of capturing local information. Standard 2D CNNs are widely applied in image processing problems while for unsqaured meshes, 1D CNN and Graph \acp{NN} \cite{zhou2020fully} are often prioritised due to the irregular structure. For more details about CNN and CAE, interested readers are referred to~\cite{rawat2017deep}.

\subsubsection{POD AE}
\label{sec:POD AE}
The combination of \ac{POD} and \ac{AE} (also known as \ac{POD} \ac{AE} or \ac{SVD} AE) was first introduced in the recent work of \cite{phillips2021autoencoder} for applications in nuclear engineering. The accuracy and efficiency of this approach has also been \rev{assessed} in urban pollution applications (e.g., \cite{quilodran2021adversarial}), especially for problems with unstructured meshes. This method consists of two steps of dimension reduction. We first apply the \ac{POD} to obtain the full set of principle components of the associated dynamical system. \rev{Using a part of the principle components as input}, a dense autoencoder with fully connected neural networks is then \rev{employed} to further reduce the problem dimension~\cite{quilodran2021adversarial}. \rev{As an important note, including
all of the PCs can involve some redundancy and noise which affects the performance of the AE. To avoid such effect, a prior \ac{POD} truncation can be performed.} 
In other words, both the input and ouput of this \ac{AE} (with Encoder $\mathcal{E}'_{\bx}$ and Decoder $\mathcal{D}'_{\bx}$) are the compressed latent vectors $\tilde{\bx}_\lambda$ associated with the \rev{\ac{POD} coefficients}, i.e.,
\begin{align}
    \tilde{\bx}_\lambda = {{\bL}_{q',\bX}}^T \bx, \quad \tilde{\bx} =  \mathcal{E}'_{\bx} (\tilde{\bx}_\lambda) \quad \textrm{while} \quad  \tilde{\bx}^r_\lambda =  \mathcal{D}'_{\bx} (\tilde{\bx}), \quad \bx^r_\textrm{POD AE} = {\bL}_{q',\bX} \hspace{1mm} \tilde{\bx}^r_\lambda 
\end{align}
where $\tilde{\bx}^r_\lambda $ and $\bx^r_\textrm{POD AE}$ denote the reconstruction of the \rev{\ac{POD} coefficients} and the reconstruction of the full physical field respectively. \rev{The prior \ac{POD} truncation parameter is denoted as $q'$.}
Since the \ac{POD} \rev{considerably reduce the size of the input vectors in AE}, applying fully connected \acp{NN} layers is computationally affordable without the concern of over-parameterization as pointed out by \cite{phillips2021autoencoder}. Furthermore, the training time will be reduced in comparison to a full \ac{CNN} \ac{AE} applied directly to the high-fidelity solutions. It is important to point out that convolutional layers can also be used in the \ac{POD} \ac{AE} approach.


\subsection{Surrogate model construction and monitoring}
\label{sec:LSTM}
Now that the \ac{ROM} is performed, we aim to construct a lower-dimensional surrogate model by understanding the evolution of the latent variables. For this purpose, we build a \ac{ML} surrogate model in the latent space, which is trained by encoded simulation data.
With the development of \ac{ML} techniques, there is an
increasing interest in using RNNs to learn the dynamics of \ac{CFD} or geoscience applications. Addressing temporal sequences
as directed graphs, RNNs manage to handle complex dynamical systems because of their ability of capturing historical dependencies through feedback loops \cite{mikolov2010recurrent}. However, training standard RNNs to solve problems with long-term temporal dependencies can be computationally difficult because the gradient of the loss function \rev{may decrease exponentially with time.} This is also known as the vanishing gradient problem \cite{hochreiter1998vanishing}. A specific type of RNN, the long-short-term-memory (LSTM) network is developed to deal with long-term temporal dependencies. In brief, different from standard \ac{RNN} units, LSTM units $C^\textrm{LSTM}_{t}$ (here $t$ denotes the time) are capable of maintaining information in memory of long periods with the help of a memory cell. Three gates, each composed
of \rev{a Sigmoid activation function  $\sigma(x) = (1/(1 + e^{-x}))$}, are used to decide when information is memorised or forgotten. The different gates and their transition functions are listed herebelow:
\begin{itemize}
    \item \textit{Forget gate} decides whether the information is going to be forgotten for the current cell unit. Here the recurrent variable $\mathbf{h}_{t-1}$ summarises all historical information and $\mathbf{x}_t$ is the current layer input,
    \begin{align}
\label{eq:forget_gate}
  f^{LSTM}_t=\sigma(\mathbf{W}_f\cdot[\mathbf{h}_{t-1},\mathbf{x}_t]+b_f) 
\end{align}

    \item \textit{Input gate} determines the new information which is going to be added with
        \begin{align}
\label{eq:tanh_layer}
\tilde{C}^{LSTM}_{t}=\tanh(\mathbf{W}_C\cdot[\mathbf{h}_{t-1},\mathbf{x}_t]+b_C),
\end{align}
\begin{align}
\label{eq:input_gate}
\mathbf{i}_{t}=\sigma(\mathbf{W}_i\cdot[\mathbf{h}_{t-1},\mathbf{x}_t]+b_i),
\end{align}
while $\tilde{C}^{LSTM}_{t}$ is multiplied by weight coefficients, leading to an update of $C^{LSTM}_{t}$,
\begin{align}
\label{eq:cell_state}
C^{LSTM}_{t}=f^{LSTM}_t\odot C^{LSTM}_{t-1}+\mathbf{i}_t\odot \tilde{C}^{LSTM}_{t},
\end{align}
where $\odot $ denotes the Hadamard product of vectors and matrices.
\item \textit{Output gate} decides the recurrent state $\mathbf{h}_t$ as a function of previous recurrent output $\mathbf{h}_{t-1}$ and the current layer input $\mathbf{x}_t$ through a Sigmoid activation function, i.e.,
\begin{align}
\label{eq:output_gate}
\mathbf{o}_{t}=\sigma(\mathbf{W}_o[\mathbf{h}_{t-1},\mathbf{x}_t]+b_o)
\end{align}
\begin{align}
\label{eq:recurrent_output}
\mathbf{h}_{t}=\mathbf{o}_t\odot \tanh(C^{LSTM}_t)
\end{align}
\end{itemize}
Here $\bW$ and $\bb$ denote \rev{the weight and the bias coefficients for different gates respectively.} Once the LSTM \ac{NN} is trained in the latent space, a low dimensional surrogate model can then be established for predicting the evolution of the dynamical system with a low computational cost.  

\section{Methodology: Generalised Latent Assimilation}
\label{sec:latent assimilation}
Latent Assimilation techniques \cite{amendola2020, peyron2021latent} have been developed for the real-time monitoring of latent surrogate models. Here we have developed a new generalised \ac{LA} approach which can incorporate observation data encoded in  a latent space different from the one of state variables. \rev{Since we aim to assimilate a dynamical system, the dependence on time $t$ is introduced for all state/observation variables in the rest of this paper.}  \\

\subsection{Variational assimilation principle}
\label{sec:VDA}
\label{sec:DA}
\noindent Data assimilation algorithms aim to improve the prediction of some physical fields (or a set of parameters) $\textbf{x}_t$ based on two sources of information: a prior forecast $\textbf{x}_{b,t}$ (also known as the background state) and an observation vector $\textbf{y}_t$. The true state which represents the theoretical value of the current state is denoted by $\textbf{x}_{\textrm{true},t}$. 
In brief, Variational DA searches for an optimal weight between $\textbf{x}_{b,t}$ and  $\textbf{y}_t$ by minimising the cost function $J$ defined as
\begin{align}
    J_t(\textbf{x})&=\frac{1}{2}(\textbf{x}-\textbf{x}_{b,t})^T\textbf{B}_t^{-1}(\textbf{x}-\textbf{x}_{b,t}) + \frac{1}{2}(\textbf{y}_t-\mathcal{H}_t(\textbf{x}))^T \textbf{R}_t^{-1} (\textbf{y}_t-\mathcal{H}_t(\textbf{x}_t)) \label{eq_3dvar}\\
   &=\frac{1}{2}\vert \vert\textbf{x}-\textbf{x}_{b,t}\vert \vert^2_{\textbf{B}_t^{-1}}+\frac{1}{2}\vert \vert\textbf{y}_t-\mathcal{H}_t(\textbf{x})\vert \vert^2_{\textbf{R}_t^{-1}} \notag
\end{align}
 where  $\mathcal{H}_t$ denotes the state-observation mapping function, and $\textbf{B}_t$ and $\textbf{R}_t$ are the error covariance matrices related to  $\textbf{x}_{b,t}$ and  $\textbf{y}_t$, i.e.,
 \begin{align}
     \textbf{B}_t = \textrm{Cov}(\epsilon_{b,t}, \epsilon_{b,t}), \quad
     \textbf{R}_t = \textrm{Cov}(\epsilon_{y,t}, \epsilon_{y,t}),
 \end{align}
 where
  \begin{align}
     \epsilon_{b,t} = \textbf{x}_{b,t} - \textbf{x}_{\textrm{true},t}, \quad
     \epsilon_{y,t} = \mathcal{H}_t(\textbf{x}_{\textrm{true},t})-\textbf{y}_t.
 \end{align}
 Since DA algorithms often deal with problems of large dimension, for the sake of simplicity, prior errors $\epsilon_b, \epsilon_y$ are often supposed to be centered Gaussian, i.e.,
 \begin{align}
     \epsilon_{b,t} \sim \mathcal{N} (0, \textbf{B}_t), \quad
     \epsilon_{y,t} \sim \mathcal{N} (0, \textbf{R}_t).
 \end{align}
Equation~\eqref{eq_3dvar}, also known as the three-dimensional variational (3D-Var) formulation, represents the general objective function of variational assimilation. Time-dependent variational assimilation (so called 4D-Var) formulation can also be reformulated into Equation~\eqref{eq_3dvar} as long as the error of the forward model is not considered. The minimisation point of equation~\eqref{eq_3dvar} is denoted as $\bx_{a,t}$, 
  \begin{align}
    \bx_{a,t} = \underset{\bx}{\argmin} \Big(J_t(\textbf{x})\Big) \label{eq:argmin},
 \end{align}
 known as the \rev{analysis} state. 
 \rev{When $\mathcal{H}_t$ is non-linear, approximate iterative methods \cite{lawless2005approximate} have been widely used to solve variational data assimilation.} To do so, one has to compute the gradient $\nabla J(\bx)$, which can be approximated by
 \begin{align}
     \nabla J(\bx) \approx 2 \bB_t^{-1}(\bx-\bx_{b,t}) - 2 \bH^T \bR_t^{-1} (\by_t-\mathcal{H}_t(\bx)). \label{eq:gradient}
 \end{align}
 In equation~\eqref{eq:gradient}, $\bH$ is obtained via a local linearization in the neighbourhood of the current vector $\bx$. The minimization of 3D-Var is often performed via quasi-Newton methods, including for instance BFGS approaches \cite{Fulton2000}, where each iteration can be written as:
  \begin{align}
     \bx_{k+1} = \bx_{k} - L_\textrm{3D-Var} \big[ \textrm{Hess}(J) (\bx_{k})\big]^{-1} \nabla J(\bx_k)
 \end{align}
Here $k$ is the current iteration, and $L_\textrm{3D-Var}>0$ is the learning rate of the descent algorithm, and
\begin{align}
\textrm{Hess}\Big(J(\bx = [x_0,...,x_{n-1}])\Big)_{i,j} = \frac{\partial^2 J}{\partial x_i \partial x_j}
\end{align}
is the Hessian matrix related to the cost function $J$. The process of the iterative minimization algorithm is summarised in Algorithm~\ref{algo:1}.

\begin{algorithm}
\begin{algorithmic}
\caption{Iterative minization of 3D-Var cost function via quasi-Newton methods }
\label{algo:1}
\State Inputs: $\bx_{b,t}, \by_t, \bB_t, \bR_t, \mathcal{H}_t$
\State parameters: $k_\textrm{max}$ , $\epsilon$ 
\State $\bx_0 = \bx_b,$ $ k = 0$\\

\While{$k<k_\textrm{max}$ \textrm{and} $\vert \vert \nabla J_t(\bx_k)\vert \vert  > \epsilon$ }
        \State $J_t(\textbf{x}_k) = \frac{1}{2}\vert \vert\textbf{x}_k-\textbf{x}_{b,t}\vert \vert^2_{\textbf{B}_t^{-1}}+\frac{1}{2}\vert \vert\textbf{y}_t-\mathcal{H}_t(\textbf{x}_k)\vert \vert^2_{\textbf{R}_t^{-1}}$ 
        \rev{\State linearize the $\mathcal{H}_t$ operator in the neighbourhood of $\bx_k$
        \State $\nabla J_t(\bx_k) \approx 2 \bB_t^{-1}(\bx_k-\bx_{b,t}) - 2 \bH^T \bR_t^{-1} (\by_t-\mathcal{H}_t(\bx_k))$}
        \State compute $\textrm{Hess}\big(J_t(\bx_k)\big)$
        \State $ \bx_{k+1} = \bx_{k} - L_\textrm{3D-Var} \big[ \textrm{Hess}(J) \bx_{k}\big]^{-1} \nabla J_t(\bx_k)$
        \State \textit{k} = \textit{k}+1
\EndWhile

output: $\bx_k$
\end{algorithmic}
\end{algorithm}

Variational assimilation algorithms could be applied to dynamical systems for improving future prediction by using a transition operator $\mathcal{M}_{t^k \rightarrow t^{k+1}}$ (from time $t^k$ to $t^{k+1}$), thus
\begin{align}
    \textbf{x}_{t^{k+1}} = \mathcal{M}_{t^k \rightarrow t^{k+1}} (\textbf{x}_{t^k }).
\end{align}
In our study, the $\mathcal{M}_{t^k \rightarrow t^{k+1}}$ operator is defined by a latent LSTM surrogate model.
 Typically in DA, the current background state is often provided by the forecasting from the previous time step, i.e.
\begin{align}
    \textbf{x}_{b,t^k} = \mathcal{M}_{t^{k-1} \rightarrow t^{k}} (\textbf{x}_{a,t^{k-1} }).
\end{align}
A more accurate reanalysis $\textbf{x}_{a,t^{k-1}}$ leads to a more reliable forecasting $\textbf{x}_{b,t^k}$. However, in practice, \rev{the perfect knowledge of} $\mathcal{M} $ is often out of reach. Recent work of \cite{Brajard2020} makes use of deep learning algorithms to improve the estimation of  $\mathcal{M}_{t^{k-1} \rightarrow t^{k}}$. From Algorithm~\ref{algo:1}, one observes that the linearization of $\mathcal{H}$ and the \rev{evaluation} of $\textrm{Hess}\big(J(\bx_k)\big)$ is necessary for variational assimilation. Since in this application, the latent variables and observations are linked via \acp{NN} functions, the linearization and the partial derivative calculation are almost infeasible due to:
\begin{itemize}
    \item the huge number of parameters in the \acp{NN};
    \item the non-differentiability of \acp{NN} functions, for instance, when using activation functions such as ReLu or LeakyReLu \cite{wang2019learning}.
\end{itemize}
Therefore, we propose the use of a smooth local surrogate function to overcome these difficulties. \\ 

\subsection{Assimilation with heterogeneous latent spaces}
\label{sec:ahls}
\begin{figure}[h!]
\centering
\includegraphics[width = 3.9in]{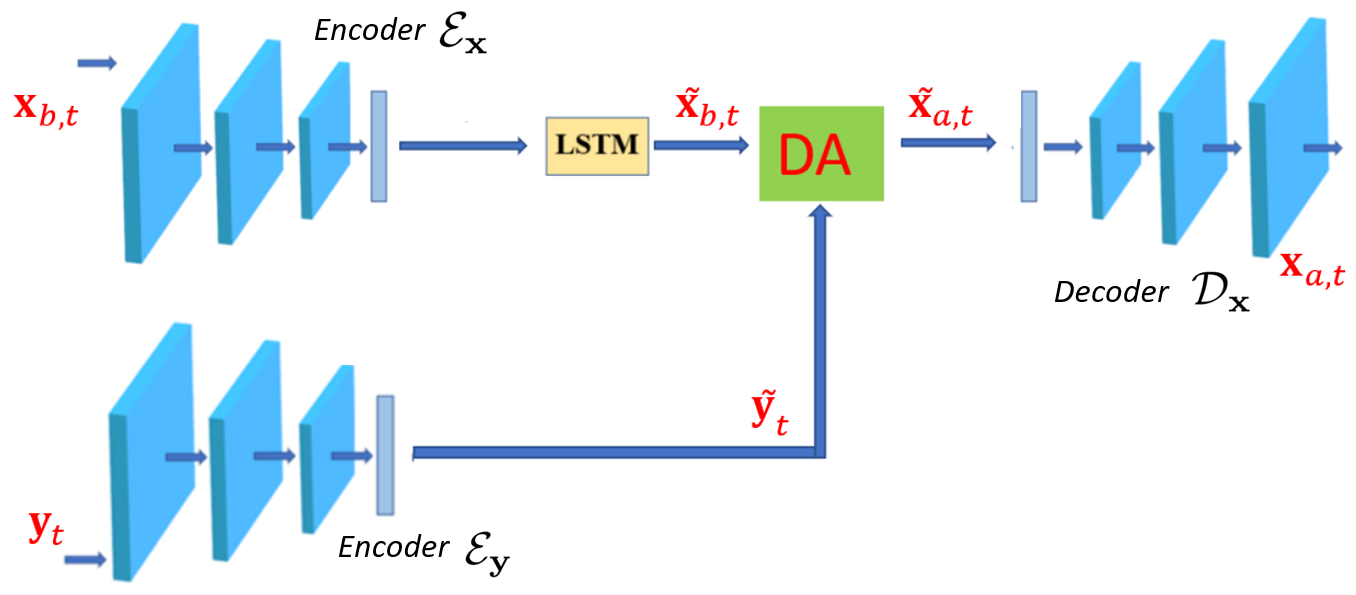} 
   \caption{Flowchart of the \ac{LA} with heterogeneous latent spaces}
   \label{fig:latentDA}
\end{figure}

\noindent Latent Assimilation techniques are introduced in the very recent work of \cite{amendola2020,peyron2021latent} where 
the DA is performed after having compressed the state and the observation data into the same latent space. In other words, it is mandatory to have the transformation operator \rev{$\tilde{\mathcal{H}}_t = \bI$} in the latent space. To fulfil this condition, \cite{amendola2020} preprocesses the observation data via a linear interpolation to the full space of the state variables. However, as mentioned in their work, this preprocessing will introduce additional errors, which may impact the assimilation accuracy. More importantly, it is almost infeasible to compress $\bx$ and $\by$ into a same latent space in a wide range of DA applications, due to, for instance:
\begin{itemize}
    \item partial observation:  only a part of the state variables are observable, usually in certain regions of the full state space;
    \item a complex $\mathcal{H}$ function in the full space: $\bx$ and $\by$ are different physical quantities (e.g., temperature vs. wind in \rev{weather prediction}, river flow vs. precipitation in hydrology).
\end{itemize}
A general latent transformation operator $\tilde{\mathcal{H}}_t$ can be formulated as
\begin{align}
    & \tilde{\mathcal{H}}_t = \mathcal{E}_\by \circ \mathcal{H}_t \circ \mathcal{D}_\bx, \quad \textrm{i.e.,} \quad \tilde{\by} = \mathcal{E}_\by \circ \mathcal{H}_t \circ \mathcal{D}_\bx \big(\tilde{\bx}\big) =  \tilde{\mathcal{H}}_t \big(\tilde{\bx} \big),\notag \\
    & \textrm{with} \quad \tilde{\by}_t = \mathcal{E}_\by (\by_t), \quad \bx_t = \mathcal{D}_\bx (\tilde{\bx}_t), \label{eq:DExy}
\end{align}
where $\mathcal{E}_\by, \mathcal{D}_\bx$ denote the encoder of the observation vectors and the decoder of the state variables respectively. A flowchart of the generalised \ac{LA} is illustrated in figure~\ref{fig:latentDA}. The cost function $\tilde{J}_t$ of general \ac{LA} problems reads

\begin{align}
    \tilde{J}_t(\tilde{\bx})&=\frac{1}{2}(\tilde{\textbf{x}}-\tilde{\textbf{x}}_{b,t})^T\tilde{\textbf{B}_t}^{-1}(\tilde{\textbf{x}}-\tilde{\textbf{x}}_{b,t}) + \frac{1}{2}(\tilde{\textbf{y}}_t-\tilde{\mathcal{H}_t}(\rev{\tilde{\textbf{x}}})^T \tilde{\textbf{R}_t}^{-1} (\tilde{\textbf{y}}_t-\tilde{\mathcal{H}}_t(\tilde{\textbf{x}})). \label{eq_latent3dvaroriginal} \\
    \tilde{\bx}_a &= \underset{\tilde{\bx}}{\argmin} \Big(\tilde{J}_t(\tilde{\bx})\Big). \label{eq:latentargmin}
\end{align}

\rev{In the rest of this paper, it is supposed that the latent error covariances $\tilde{\bB}_t =  \bB, \tilde{\bR}_t =  \bR$ are time invariant.}
\subsection{\rev{Polynomial regression for surrogate transformation function}}
Despite the fact that traditional variational DA approaches can deal with complex $\mathcal{H}$ functions, it is almost impossible to perform descent methods for Algorithm~\ref{algo:1} because of the drawbacks described at the end of Section~\ref{sec:VDA}.
Our idea consists of building a local smooth and differentiable surrogate function $\tilde{\mathcal{H}}_t^p $ such  that
\begin{align}
    \tilde{\mathcal{H}}_t^p (\tilde{\bx_t^s}) \approx \tilde{\mathcal{H}}_t (\tilde{\bx_t^s}) \quad \textrm{for} \quad  \textrm{$\tilde{\bx_t^s}$ in a neighbourhood of $\tilde{\bx}_{b,t}$}.
\end{align}
It is important to note that the computation of $\tilde{\mathcal{H}}^p$ will also depend on the value of the latent variable $\tilde{\bx}$.
The approximate cost function can then be written as
\begin{align}
    \tilde{J}_t^p(\rev{\tilde{\textbf{x}}})&=\frac{1}{2}(\tilde{\textbf{x}}-\tilde{\textbf{x}}_{b,t})^T\tilde{\textbf{B}}^{-1}(\tilde{\textbf{x}}-\tilde{\textbf{x}}_b) + \frac{1}{2}(\tilde{\textbf{y}}_t-\rev{\tilde{\mathcal{H}_t^p}(\tilde{\textbf{x}}))^T} \tilde{\textbf{R}}^{-1} (\tilde{\textbf{y}}_t-\tilde{\mathcal{H}_t^p}(\tilde{\textbf{x}})). \label{eq_latent3dvar}
\end{align}
The way of computing the surrogate function makes crucial impact on both the accuracy and the computational cost of \ac{DA} since the $\tilde{\mathcal{H}}$ function may vary a lot with time for chaotic dynamical systems. From now, we denote  $\tilde{\mathcal{H}}_t$ and  $\tilde{\mathcal{H}}_t^p$, the latent transformation function at time $t$ and the associated surrogate function. For time variant $\tilde{\mathcal{H}}_t$ and $\bx_t$, the computation of $\tilde{\mathcal{H}}_t^p$ must be performed online. Thus the choice of local surrogate modelling approach should be a tradeoff of approximation accuracy and computational time. As mentioned in the Introduction of this paper, the idea of computing local surrogate model has been developed in the field of interpretable AI. Linear regression (including Lasso, Ridge) and simple \ac{ML} models such as DT are prioritised for the sake of interpretability (e.g., \cite{ribeiro2016should}). In this study, the local surrogate function is built via polynomial regression since our main criteria \rev{are smoothness and differentiability}. Compared to other approaches, employing \ac{PR} in \ac{LA} has several advantages in terms of smoothness and computing efficiency.

\begin{figure}[h!]
\centering
\includegraphics[width = 4.0in]{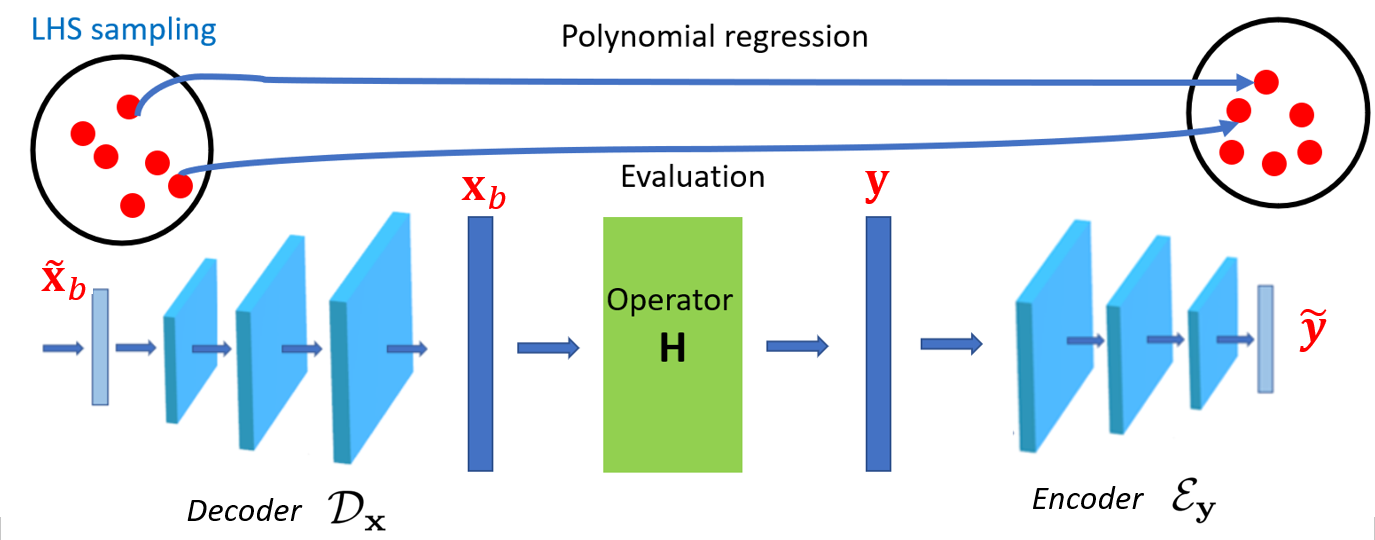} 
   \caption{Flowchart of the polynomial-based local surrogate model in Latent Assimilation.}
   \label{fig:PR}
\end{figure}

To perform the local \ac{PR}, we rely on local training datasets $\{ \tilde{\bx}_{b,t}^q \}_{q = 1.. n_s}$ generated randomly around the current background state $\tilde{\bx}_{b,t}$ since the true state is out of reach. The sampling is performed using \ac{LHS} to efficiently cover the local neighbourhood homogeneously~\cite{tang1993orthogonal}. Other sampling techniques, such as Gaussian perturbation, can also be considered regarding the prior knowledge of the dynamical system.  We then fit the output of the transformation operator by a local polynomial function,

\begin{align}
    \rev{\tilde{\mathcal{H}}_t^p = \underset{{p \in P(d_p)}}{\argmin} \big( \sum_{q=1}^{n_s} \lvert\lvert p({\bx}_{b,t}^q) - \mathcal{H}_t ({\bx}_{b,t}^q)\lvert\lvert^2_2\big)^{1/2}},
\end{align}
\rev{where $P(d_p)$ represents the set of polynomial functions of degree $d_p$.}
We then evaluate the $\tilde{\mathcal{H}}_t$ function to generate the learning targets of local \ac{PR} as shown in figure~\ref{fig:PR}. The pipeline of the \ac{LA} algorithms for dynamical models is summerised in Algorithm~\ref{algo:2}, where $\tilde{\mathcal{M}}$ denotes the forward operator in the latent space. In the context of this paper, $\tilde{\mathcal{M}}$ is the latent LSTM surrogate model. When using a \rev{sequence-to-sequence} prediction, the forecasting model can be accelerated in the sense that a \rev{sequence} of future background states can be predicted by one evaluation of LSTM. The \ac{PR} degree, the sampling range and the sampling size are denoted as $d_p, r_s$ and $n_s$ respectively. These parameters affect considerably the performance of Generalised LA. Their values should be chosen carefully as shown later in Section~\ref{sec:numerical validation}. 

\begin{algorithm}[h!]
\caption{Generalised \ac{LA} with local polynomial surrogate function}
\begin{algorithmic}[1]
\State Inputs: $\rev{\tilde{\bx}_{b,0}}, \{\by_t\}, \rev{\mathcal{E}_y}, \tilde{\bB}, \tilde{\bR}, \tilde{\mathcal{H}}, \tilde{\mathcal{M}}$
\State paramters: $d_p$, $r_s$, $n_s$, $T$
\State $\tilde{\bx}_0 = \tilde{\bx}_b,$ $ k = 0$
\For {$t$ = $0$ to $T$}  
 \State   $\tilde{\bx}_{b,t} = \tilde{\mathcal{M}} \big( \tilde{\bx}_{t-1}, \tilde{\bx}_{t-2}, \tilde{\bx}_{t-3}, ...\big)$
  \If{$\by_t$ is available}
  \State \rev{$\tilde{\by}_t = \mathcal{E}_y (\by_t)$}
 \State$\{ \tilde{\bx}_{b,t}^q \}_{q = 1.. n_s} = \textrm{LHS}_{\{d^{p},r_s,n_s\}} (\tilde{\bx}_{b,t})$
  \For {$q$ = $0$ to $n_s$}
 \State $\tilde{\by}_t^q = \tilde{\mathcal{H}} (\tilde{\bx}_{b,t}^q) $
 \EndFor
 \State $\tilde{\mathcal{H}}_t^p = \textrm{PR}_\textrm{train} \big( \textrm{input:} \{ \tilde{\bx}_{b,t}^q \}, \textrm{output:}\{ \tilde{\by}_t^q \}, {q = 1.. n_s}\big)$
 \State \textit{optional:}  $\{ \tilde{\bx}_\textrm{test}^q \}_{q = 1.. n_s} = \textrm{LHS Sampling}_{\{d^{p},r_s,n_s\}} (\tilde{\bx}_{b,t})$\\
\vspace{0.7mm}
 \State \textit{optional:} $\epsilon^p_{r-rmse} = \sqrt{\frac{1}{n_s} \sum_{q}^{n_s} \big( \vert \vert\tilde{\mathcal{H}_t} (\tilde{\bx}_{test}^q)-\tilde{\mathcal{H}}_t^p (\tilde{\bx}_\textrm{test}^q)\vert \vert^2 \big/ \vert \vert\tilde{\mathcal{H}_t} (\tilde{\bx}_{test}^q)\vert \vert^2 \big)}$ 
\State $\tilde{\bx}_{a,t}= \underset{\tilde{\bx}}{\argmin} \Big( \frac{1}{2}\vert \vert\tilde{\bx}-\tilde{\bx}_{b,t}\vert \vert^2_{\tilde{\textbf{B}}^{-1}}+\frac{1}{2}\vert \vert \tilde{\textbf{y}}_t-\tilde{\mathcal{H}}_t^p(\tilde{\bx})\vert \vert^2_{\tilde{\textbf{R}}^{-1}} \Big)$ 
 \State $\tilde{\bx}_t = \tilde{\bx}_{a,t}$ 
 
\Else
\State $\tilde{\bx}_t = \tilde{\bx}_{b,t}$
 \EndIf
\EndFor
\end{algorithmic}
\label{algo:2}
\end{algorithm}

\subsection{\rev{Theoretical analysis of the loss function}}
Since the latent variational assimilation is completely determined by its cost function $J_t^p(\tilde{\bx})$, we aim to provide a theoretical upper bound for the expected absolute and relative approximation error evaluated on the true state, i.e.,
\begin{align}
    \mathbb{E} \big( J_t^p(\tilde{\bx}_{\textrm{true},t}) - J_t(\tilde{\bx}_{\textrm{true},t}) \big) \quad \textrm{and} \quad \frac{\mathbb{E} \big( J_t^p(\tilde{\bx}_{\textrm{true},t}) - J_t(\tilde{\bx}_{\textrm{true},t}) \big)}{\mathbb{E} \big(J_t(\tilde{\bx}_{\textrm{true},t}))}.
\end{align}
In fact, the difference between $J_t(\tilde{\bx})$ and $J_t^p(\tilde{\bx})$ for any point $\tilde{\bx}$ in the space can be bounded as
\begin{align}
    J_t^p(\tilde{\bx}) &= \frac{1}{2} \Big( \vert \vert\tilde{\bx}-\tilde{\bx}_{b,t}\vert \vert^2_{\tilde{\textbf{B}}^{-1}}+\vert \vert\tilde{\textbf{y}}_t-\tilde{\mathcal{H}}_t(\tilde{\bx}) + \tilde{\mathcal{H}}_t(\tilde{\bx}) - \tilde{\mathcal{H}}_t^p(\tilde{\bx})\vert \vert^2_{\tilde{\textbf{R}}^{-1}} \Big) \\
    & \leq \frac{1}{2} \Big( \vert \vert\tilde{\bx}-\tilde{\bx}_{b,t}\vert \vert^2_{\tilde{\textbf{B}}^{-1}}+\vert \vert\tilde{\textbf{y}}_t-\tilde{\mathcal{H}}_t(\tilde{\bx})\vert \vert^2_{\tilde{\textbf{R}}^{-1}} +\vert \vert \tilde{\mathcal{H}}_t(\tilde{\bx})- \tilde{\mathcal{H}}_t^p(\tilde{\bx})\vert \vert^2_{\tilde{\textbf{R}}^{-1}} \notag \\
    &+ 2\vert \vert\tilde{\textbf{y}}_t-\tilde{\mathcal{H}}_t(\tilde{\bx})\vert \vert_{\tilde{\textbf{R}}^{-1}} \cdot \vert \vert \tilde{\mathcal{H}}_t(\tilde{\bx})- \tilde{\mathcal{H}}_t^p(\tilde{\bx})\vert \vert_{\tilde{\textbf{R}}^{-1}}\Big)\\
    & \leq \frac{1}{2} \Big( J_t(\tilde{\bx}) + \vert \vert \tilde{\mathcal{H}}_t(\tilde{\bx})- \tilde{\mathcal{H}}_t^p(\tilde{\bx})\vert \vert^2_{\tilde{\textbf{R}}^{-1}}\Big) + \vert \vert\tilde{\textbf{y}}_t-\tilde{\mathcal{H}}_t(\tilde{\bx})\vert \vert_{\tilde{\textbf{R}}^{-1}} \cdot \vert \vert \tilde{\mathcal{H}}_t(\tilde{\bx})- \tilde{\mathcal{H}}_t^p(\tilde{\bx})\vert \vert_{\tilde{\textbf{R}}^{-1}}. \label{eq:diff Jtp}
\end{align}
We are interested in the expectation value of the loss function evaluated on the true state, i.e., $\mathbb{E}(J_t^p(\tilde{\bx}_{\textrm{true},t}))$. 
Following equation.~\eqref{eq:diff Jtp}, 
\begin{align}
    \mathbb{E} \big( J_t^p(\tilde{\bx}_{\textrm{true},t}) \big) & \leq  \mathbb{E} \big( J_t(\tilde{\bx}_{\textrm{true},t}) \big)+ \frac{1}{2} \mathbb{E} \big( \vert \vert \tilde{\mathcal{H}}_t(\tilde{\bx}_{\textrm{true},t})- \tilde{\mathcal{H}}_t^p(\tilde{\bx}_{\textrm{true},t})\vert \vert^2_{\tilde{\textbf{R}}^{-1}} \big)\notag \\
    & + \mathbb{E} \big( \vert \vert\tilde{\textbf{y}}_t-\tilde{\mathcal{H}}_t(\tilde{\bx}_{\textrm{true},t})\vert \vert_{\tilde{\textbf{R}}^{-1}} \cdot \vert \vert \tilde{\mathcal{H}}_t(\tilde{\bx}_{\textrm{true},t})- \tilde{\mathcal{H}}_t^p(\tilde{\bx}_{\textrm{true},t})\vert \vert_{\tilde{\textbf{R}}^{-1}} \big) \label{eq:exp Jtp}.
\end{align}
\rev{In the case of ideal data assimilation}, both background and observation prior errors follow a centred Gaussian distribution, i.e., 
\begin{align}
    \tilde{\bx}_{b,t} - \tilde{\bx}_{\textrm{true},t} \sim \mathcal{N}(0, \tilde{\bB}), \quad \tilde{\bf{y}}_t-\tilde{\mathcal{H}}_t(\tilde{\bx}_{\textrm{true},t}) \sim ( 0, \tilde{\bR})
\end{align}
As a consequence, 
\begin{align}
\sqrt{\tilde{\bB}^{-1}} (\tilde{\bx}_{b,t} - \tilde{\bx}_{\textrm{true},t}) \sim \mathcal{N}(0, \bI_{\textrm{dim}(\tilde{\bx})}), \quad \sqrt{\tilde{\bR}^{-1}} (\tilde{\bf{y}}_t-\tilde{\mathcal{H}}_t(\tilde{\bx}_{\textrm{true},t}) )\sim \mathcal{N}( 0, \mathbf{I}_{\textrm{dim}(\tilde{\by})}).
\end{align}
Here we remind that by definition, $\tilde{\bB}$ and $\tilde{\bR}$ are real constant symmetric positive definite matrices thus $\sqrt{\tilde{\bB}^{-1}} $ and $\sqrt{\tilde{\bR}^{-1}} $ are well-defined. 

\begin{align}
    \mathbb{E}(\vert \vert\tilde{\bx}_{\textrm{true},t}-\tilde{\bx}_{b,t}\vert \vert^2_{\tilde{\textbf{B}}^{-1}}) &= \mathbb{E}\Bigg((\tilde{\bx}_{\textrm{true},t}-\tilde{\bx}_{b,t})^T \tilde{\textbf{B}}^{-1} (\tilde{\bx}_{\textrm{true},t}-\tilde{\bx}_{b,t}) \Bigg)\\
    & = \mathbb{E}\Bigg(\Big( \sqrt{\tilde{\bB}^{-1}} (\tilde{\bx}_{b,t} - \tilde{\bx}_{\textrm{true},t}) \Big)^T \cdot  \Big( \sqrt{\tilde{\bB}^{-1}} (\tilde{\bx}_{b,t} - \tilde{\bx}_{\textrm{true},t}) \Big) \Bigg)\\
    & =  \mathbb{E}\Bigg(\vert \vert\sqrt{\tilde{\bB}^{-1}} (\tilde{\bx}_{b,t} - \tilde{\bx}_{\textrm{true},t})\vert \vert^2_2 \Bigg)\\
    & = \textrm{dim}(\tilde{\bx})
\end{align}
For the same reason, $ \mathbb{E}(\vert \vert\tilde{\textbf{y}}_t-\tilde{\mathcal{H}}_t(\tilde{\bx})\vert \vert^2_{\tilde{\textbf{R}}^{-1}}) = \textrm{dim}(\tilde{\by}_t)$. One can then deduce
\begin{align}
    \mathbb{E} \big( J_t (\tilde{\bx}_{\textrm{true},t}) \big) = \textrm{dim}(\tilde{\bx}_t) + \textrm{dim}(\tilde{\by}_t).
\end{align}
A similar reasoning via Mahalanobis norm can be found in the work of \cite{Talagrand1998}. \\

Now we focus on the other terms of equation~\eqref{eq:exp Jtp}. In fact, the  observation error $  \vert \vert\tilde{\textbf{y}}_t-\tilde{\mathcal{H}}_t(\tilde{\bx}_{\textrm{true},t})\vert \vert_{\tilde{\textbf{R}}^{-1}}$ is only related to instrument noises or representation error if the encoder error can be neglected. On the other hand, the approximation error $\vert \vert \tilde{\mathcal{H}}_t(\tilde{\bx})- \tilde{\mathcal{H}}_t^p(\tilde{\bx})\vert \vert_{\tilde{\textbf{R}}^{-1}}$ is only related to polynomial regression where the real observation vector $\by$ is not involved. Therefore, we can suppose that $  \vert \vert\tilde{\textbf{y}}_t-\tilde{\mathcal{H}}_t(\tilde{\bx}_{\textrm{true},t})\vert \vert_{\tilde{\textbf{R}}^{-1}}$ is uncorrelated to $\vert \vert \tilde{\mathcal{H}}_t(\tilde{\bx})- \tilde{\mathcal{H}}_t^p(\tilde{\bx})\vert \vert_{\tilde{\textbf{R}}^{-1}}$ . This assumption will be proved numerically in experiments. One can further deduce that,

\begin{align}
    & \mathbb{E} \big( \vert \vert\tilde{\textbf{y}}_t-\tilde{\mathcal{H}}_t(\tilde{\bx}_{\textrm{true},t})\vert \vert_{\tilde{\textbf{R}}^{-1}} \cdot \vert \vert \tilde{\mathcal{H}}_t(\tilde{\bx}_{\textrm{true},t})- \tilde{\mathcal{H}}_t^p(\tilde{\bx}_{\textrm{true},t})\vert \vert_{\tilde{\textbf{R}}^{-1}} \big) \notag \\
    &= \mathbb{E} \big( \vert \vert\tilde{\textbf{y}}_t-\tilde{\mathcal{H}}_t(\tilde{\bx}_{\textrm{true},t})\vert \vert_{\tilde{\textbf{R}}^{-1}} \big) \cdot \mathbb{E} \big( \vert \vert \tilde{\mathcal{H}}_t(\tilde{\bx}_{\textrm{true},t})- \tilde{\mathcal{H}}_t^p(\tilde{\bx}_{\textrm{true},t})\vert \vert_{\tilde{\textbf{R}}^{-1}} \big) = 0. \label{eq:uncorr}
\end{align}
Now we only need to bound the polynomial regression error. For this, we rely on the recent theoretical results in the work of \cite{emschwiller2020neural}, which proves that for learning a teacher \acp{NN} via polynomial regression,
\begin{align}
    N^{*} = d^{O (L/\epsilon^{*})^L} \quad \textrm{for the ReLU activation function},
\end{align}
where $N^{*} $ is the required number of samples in the training dataset, $d$ is the input dimension, $L$ is the number of \acp{NN} layers and $\epsilon^{*}$ is the relative target prediction error (i.e., in our case $\epsilon = \Big( \vert \vert \tilde{\mathcal{H}}_t(\tilde{\bx})- \tilde{\mathcal{H}}_t^p(\tilde{\bx})\vert \vert_{2} / \vert \vert \tilde{\mathcal{H}}_t(\tilde{\bx})\vert \vert_{2} \Big) \leq \epsilon^{*} $). Since we are looking for a bound of the regression error $\epsilon$,
\begin{align}
     & N^{*} = d^{\big(  c  (L/\epsilon^{*})^L \big)} \quad \textrm{where} \quad \textrm{$c$ is a real constant}\\
    \Leftrightarrow \quad & \log_d N^{*} = c  (L/\epsilon^{*})^L\\
    \Leftrightarrow \quad & \Big({\frac{\log_d N^{*}}{c}}\Big)^{1/L} = L/\epsilon^{*} \\
    \Leftrightarrow \quad & \epsilon \leq \epsilon^{*} = L  \Big( \frac{c}{\log_d N^{*}} \Big)^{1/L} \\
    \Leftrightarrow \quad & \vert \vert \tilde{\mathcal{H}}_t(\tilde{\bx})- \tilde{\mathcal{H}}_t^p(\tilde{\bx})\vert \vert_{2} \leq L  \Big( \frac{c}{\log_d N^{*}} \Big)^{1/L} \vert \vert \tilde{\mathcal{H}}_t(\tilde{\bx})\vert \vert_{2}. \label{eq:norm2}
\end{align}
Now that we have a \rev{relative bound} of the polynomial prediction error in the $L^2$ norm, we want to extend this boundary to the matrix norm ${\vert \vert.\vert \vert}_{\tilde{\textbf{R}}^{-1}}$. For this we use a general algebraic result:
\begin{align}
    & \forall a \in \mathbb{R}^{\textrm{dim}(a)}, \quad \bC_{p,d} \in \mathbb{R}^{\textrm{dim}(a) \times \textrm{dim}(a)} \quad \textrm{ is a symmetric positive definite matrix then} \notag\\
    &\sqrt{\lambda_{\textrm{min}}}  \vert \vert a \vert \vert_2 \leq  \vert \vert a \vert \vert_{\bC_{p,d}} \leq \sqrt{\lambda_{\textrm{max}}}  \vert \vert a \vert \vert_2 \label{eq:eigenv}
\end{align}
where $\lambda_{\textrm{min}}, \lambda_{\textrm{max}}$ represent the smallest and the largest eigenvalues of $\bC_{p,d}$ respectively. Since $\bC_{p,d}$ is positive definite, $0 < \lambda_{\textrm{min}} \leq \lambda_{\textrm{max}}$. We denote $ 0 < {\lambda^{\tilde{\bR}}}_{\textrm{dim}(\tilde{\by})} \leq ... \leq {\lambda^{\tilde{\bR}}}_{1} $ the eigenvalues of $\tilde{\bR}$. Thus the eigenvalues of ${\tilde{\bR}}^{-1}$ are $ 0 < 1/{\lambda^{\tilde{\bR}}}_{1} \leq ... \leq 1/{\lambda^{\tilde{\bR}}}_{\textrm{dim}(\tilde{\by})}$. Following the result of Equation~\eqref{eq:eigenv},
\begin{align}
    &\vert \vert \tilde{\mathcal{H}}_t(\tilde{\bx})\vert \vert_{2} \leq \sqrt{\lambda^{\tilde{\bR}}_1} \vert \vert \tilde{\mathcal{H}}_t(\tilde{\bx})\vert \vert_{\tilde{\textbf{R}}^{-1}} \quad \textrm{and} \notag\\
    &\vert \vert \tilde{\mathcal{H}}_t(\tilde{\bx})- \tilde{\mathcal{H}}_t^p(\tilde{\bx})\vert \vert_{2} \geq  \sqrt{\lambda^{\tilde{\bR}}_{\textrm{dim}(\tilde{\by})}} \vert \vert \tilde{\mathcal{H}}_t(\tilde{\bx})- \tilde{\mathcal{H}}_t^p(\tilde{\bx})\vert \vert_{\tilde{\textbf{R}}^{-1}}.
\end{align}
Therefore, we can deduce from Equation~\eqref{eq:norm2} that
\begin{align}
    \vert \vert \tilde{\mathcal{H}}_t(\tilde{\bx})- \tilde{\mathcal{H}}_t^p(\tilde{\bx})\vert \vert_{\tilde{\textbf{R}}^{-1}} \leq \sqrt{\lambda^{\tilde{\bR}}_1/\lambda^{\tilde{\bR}}_{\textrm{dim}(\tilde{\by})}}  L  \Big( \frac{c}{\log_d N^{*}} \Big)^{1/L}  \vert \vert \tilde{\mathcal{H}}_t(\tilde{\bx})\vert \vert_{\tilde{\textbf{R}}^{-1}}.
\end{align}
Thus, 
\begin{align}
     &\mathbb{E} \big( \vert \vert \tilde{\mathcal{H}}_t(\tilde{\bx}_{\textrm{true},t})- \tilde{\mathcal{H}}_t^p(\tilde{\bx}_{\textrm{true},t})\vert \vert^2_{\tilde{\textbf{R}}^{-1}} \big) \notag \\
     & = \textrm{cond}(\bR)  L^2  \Big( \frac{c}{\log_d N^{*}} \Big)^{2/L}  \mathbb{E} \big( \vert \vert \tilde{\mathcal{H}}_t(\tilde{\bx}_{\textrm{true},t})\vert \vert^2_{\tilde{\textbf{R}}^{-1}}), \label{eq:condR}
\end{align}
where $\textrm{cond}(\bR) = \lambda^{\tilde{\bR}}_1/\lambda^{\tilde{\bR}}_{\textrm{dim}(\tilde{\by})}$ is the condition number of the $\bR$ matrix. Combining equation~\eqref{eq:exp Jtp},~\eqref{eq:uncorr} and~\eqref{eq:condR}, 
\begin{align}
   & \mathbb{E} \big( J_t^p(\tilde{\bx}_{\textrm{true},t}) \big) \notag \\
   & \leq  \mathbb{E} \big( J_t(\tilde{\bx}_{\textrm{true},t}) \big)+ \frac{1}{2} \textrm{cond}(\bR)  L^2  \Big( \frac{c}{\log_d N^{*}} \Big)^{2/L}  \mathbb{E} \big( \vert \vert \tilde{\mathcal{H}}_t(\tilde{\bx}_{\textrm{true},t})\vert \vert^2_{\tilde{\textbf{R}}^{-1}}) \notag \\
    & = \textrm{dim}(\tilde{\bx}_t) + \textrm{dim}(\tilde{\by}_t) + \frac{1}{2} \textrm{cond}(\bR)  L^2  \Big( \frac{c}{\log_d N^{*}} \Big)^{2/L}  \mathbb{E} \big( \vert \vert \tilde{\mathcal{H}}_t(\tilde{\bx}_{\textrm{true},t})\vert \vert^2_{\tilde{\textbf{R}}^{-1}}).
\end{align}
Therefore we have an upper bound of $\mathbb{E} \big( J_t^p(\tilde{\bx}_{\textrm{true},t}) \big) $ and $\mathbb{E} \big( J_t^p(\tilde{\bx}_{\textrm{true},t}) \big) - \mathbb{E} \big( J_t(\tilde{\bx}_{\textrm{true},t}) \big)$ which doesn't depend on the local polynomial surrogate model $\tilde{\mathcal{H}}^p_t $. An upper bound for the relative error can also be found, i.e.,
\begin{align}
    \frac{\mathbb{E} \big( J_t^p(\tilde{\bx}_{\textrm{true},t}) - J_t(\tilde{\bx}_{\textrm{true},t}) \big)}{\mathbb{E} \big(J_t(\tilde{\bx}_{\textrm{true},t}))} \leq  \frac{\textrm{cond}(\bR)  L^2  \Big( \frac{c}{\log_d N^{*}} \Big)^{2/L}  \mathbb{E} \big( \vert \vert \tilde{\mathcal{H}}_t(\tilde{\bx}_{\textrm{true},t})\vert \vert^2_{\tilde{\textbf{R}}^{-1}})}{2  (\textrm{dim}(\tilde{\bx}) + \textrm{dim}(\tilde{\by}))}. 
\end{align}
Furthermore, in the case where the target \acp{NN} is fixed and we have infinite local training data for the polynomial surrogate model,
\begin{align}
    \mathbb{E} \big( J_t^p(\tilde{\bx}_{\textrm{true},t}) - J_t(\tilde{\bx}_{\textrm{true},t}) \big) \xrightarrow{N^{*} \rightarrow +\infty} 0.
\end{align}
This result obtained is consistent with the Stone–Weierstrass theorem which reveals the fact that every continuous function defined on a closed interval can be approximated as closely as desired by a polynomial function \cite{holladay1957note}.

\section{Results: \ac{ROM} and \ac{RNN} approaches}
In this section, we describe the test case of an oil-water two-phase flow \ac{CFD} simulation, used for numerical comparison of different \ac{ML} surrogate models and \ac{LA} approaches. 
\subsection{CFD modelling}
\label{sec:CFD}

\begin{figure}[h!]
\centering
\includegraphics[width = 3.6in]{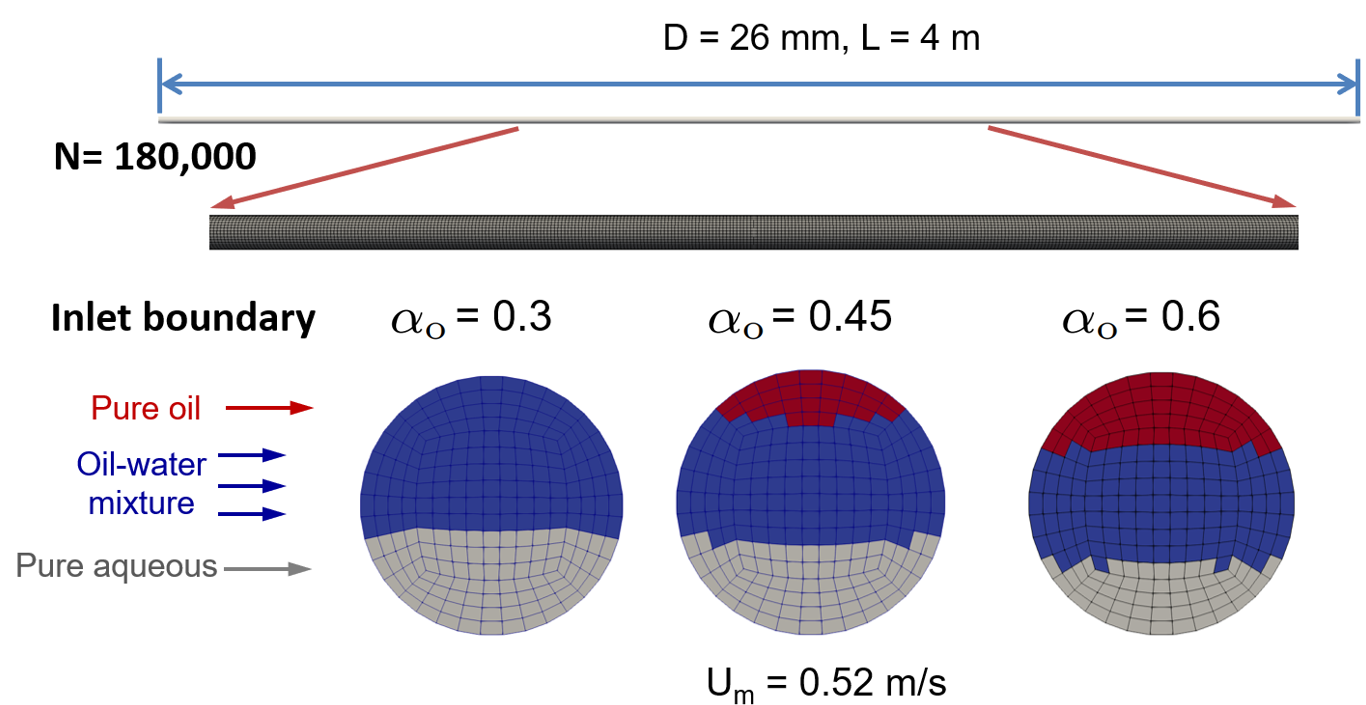}
   \caption{Dimension and parameters of the pipe and the two-phase flow}
   \label{fig:pipe}
\end{figure}

Liquid-liquid two-phase flows are widely encountered in many industrial sectors, including petroleum, chemical and biochemical engineering, food technology, pharmaceutics, and so on. In crude-oil pipelines or oil recovery equipment, both dispersed and separated oil-water flows can be observed, and the transition between these flow patterns can impact the operating cost and safety. Therefore, fundamental understanding of the oil-water flow behavior in pipelines has been tackled for a long term with various efforts from theoretical, experimental, and \rev{simulating perspectives}. However, it is not fully solved yet due to the complexity of the multiphase flow characteristics. Even for a very simple case of the separating process of oil droplets in water in a horizontal pipeline, it is still challenging to predict the separation length and layer height distribution. Although there are a lot of experimental data, the prediction of such kind of flow regime transition is still poor due to the limited understanding of the underlying physics.\\

 The experiment in this study is conducted in the flow rig developed by \cite{voulgaropoulos2017}. The flow pipe consists of a front and a back leg with equal length of 4 m and a uniform diameter of 26 mm as shown in figure~\ref{fig:pipe}. The two legs are connected by a U-bend. Measurements are conducted in the front leg only, and High-speed imaging, \rev{combined with Particle Image Velocimetry and Laser Induced Fluorescence} experiments are carried out to study the drop evolution, velocity profiles and flow patterns. \rev{As shown in table~\ref{Table2}, the two test cases explored in this work have initial mixture velocity of 0.52 m/s and 1.04 m/s respectively. The average oil inlet volume fraction of both simulations is set to 30\%.  The first simulation (i.e., the one with $U_m = 0.52 m/s$) is used to train the surrogate model while the second one is used latter to test the performance of \acp{ROM}. The simulations are validated against experimental data of the concentration profiles and layer heights.} The simulations adopt the same physical properties and operating parameters as those in the experiment. The related parameters are shown in Table 1 and Table 2. 

\begin{table}
    \centering
    \caption{Physical properties of the experimental system}
    \begin{tabular}{ccccc}
        $\begin{array}{ccccc}
        \hline \text { Liquid } & \text { Phase } & \begin{array}{c}
        \rho \\
        \left( \text{kg} \; \text{m} ^{-3}\right)
        \end{array} & \begin{array}{c}
        \mu \\
        ( \text{Pa} \; \text{s} )
        \end{array} & \begin{array}{c}
        \sigma \\
        \left( \text{N} \; \text{m} ^{-1}\right)
        \end{array} \\
        \hline \text { Water } & \text { Aqueous } & 998 & 0.89 \times 10^{-3}  &  \multirow{2}{*}{$\sim 0.0329$}\\  
        \text { Exxsol D140 } & \text { Organic } & 828 & 5.5 \times 10^{-3} &   \\
        \hline
        \end{array}$
    \end{tabular}
    \label{tab:Table 1}
\end{table}

\begin{table}
    \centering
\rev{   \caption{Operating parameters of the experiment}
    \label{Table2}
    \begin{tabular}{cc}
        $\begin{array}{cccccc}
       \hline \alpha_\mathrm{o} & h_\mathrm{ C0 }^{+}=h_\mathrm{ C0 } / D & h_\mathrm{O0}{ }^{+}=h_\mathrm{O0} / D & h_\mathrm{P0}{ }^{+}=h_\mathrm{ P0} / D & d_{320}\;\mathrm{( mm )} & U_\mathrm{ m } (\text{m s} ^{-1})\\
        \hline 
        0.3 & 0.405 & 0.997 & 0.76 & 3.41 & 0.52\\
        \hline
        0.3 & 0.189 & 0.997 & 0.92 & 1.27 & 1.04\\
        \hline
        \end{array}$
    \end{tabular}}
\end{table}

\begin{figure}[h!]
\centering
\includegraphics[width = 5.4in]{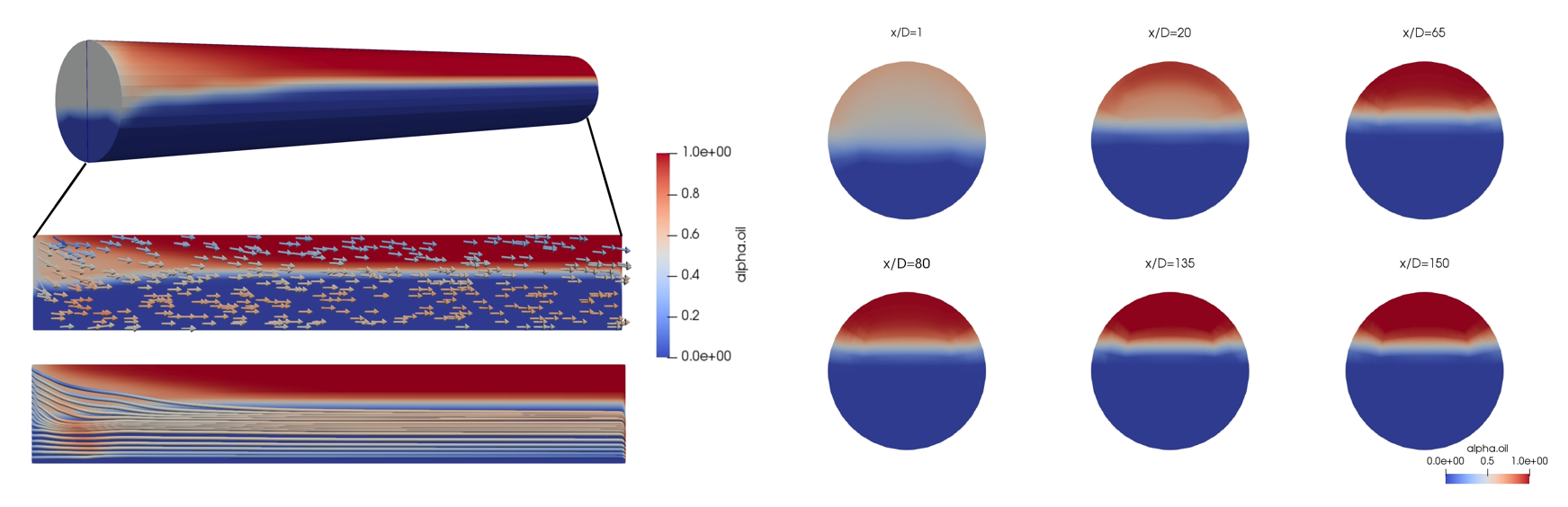}
   \caption{CFD modelling of the two-phase flow}
   \label{fig:CFDlayers}
\end{figure}
The \ac{CFD} simulation (as illustrated in figure \ref{fig:CFDlayers}) aims to study \rev{the flow separation characteristics}. The two-phase flow of silicone oil and water in a pipe with \rev{a length of 4m and a diameter of 26mm} is studied. Eulerian-Eulerian simulations are performed through the opensource \ac{CFD} platform of OpenFOAM (version 8.0), and population balance models~\cite{Kumar96} are used to model the droplet size and coalescence behaviour. The governing equations of the Eulerian framework are given as below:
\begin{equation}
	\frac{\partial}{\partial t}\left(\alpha_\mathrm{k} \rho_\mathrm{k}\right) + \nabla \cdot \left(\alpha_\mathrm{k} \rho_\mathrm{k} \textbf{\emph{U}}_\mathrm{k}\right) =0, 
\end{equation}
\begin{equation}
	\frac{\partial}{\partial t}\left(\alpha_\mathrm{k} \rho_\mathrm{k} \boldsymbol{U}_\mathrm{k}\right)+\nabla \cdot\left(\alpha_\mathrm{k} \rho_\mathrm{k} \boldsymbol{U}_\mathrm{k} \boldsymbol{U}_\mathrm{k}\right)=-\alpha_\mathrm{k} \nabla p+\nabla \cdot\left(\alpha_\mathrm{k} \boldsymbol{\tau}_\mathrm{k}\right)+\alpha_\mathrm{k} \rho_\mathrm{k} \boldsymbol{g}+\boldsymbol{M}_\mathrm{k},
\end{equation}
\noindent where the subscript of k represents the phases of water and oil respectively, and $\boldsymbol \tau$ is the stress tensor expressed as
\begin{equation}
    \boldsymbol{\tau} _\mathrm{k}=\mu_\mathrm{ eff }\left[\nabla \boldsymbol{U} _\mathrm{k}+\left(\nabla \boldsymbol{U} _\mathrm{k}\right)^\mathrm{ T }-\frac{2}{3}\left(\nabla \cdot \boldsymbol{U} _\mathrm{k}\right) I \right].
\end{equation}
A structured mesh with 180000 nodes is generated by the utility of blockMesh, and the volume concentration at the inlet boundary \rev{is prescribed by the patch manipulation (the utility of \textit{createPatch} in  OpenFOAM.)}. In all cases, the mixture $k-\epsilon$ model and wall functions are used to model turbulence equations. In order to obtain a steady flow pattern, the flow time \rev{is set to 10 s}. The time step is 0.005 s for all the cases, \rev{which ensures the convergence} at the current mesh resolution. \rev{The running time is 40 hours on a four-nodes parallel computing mode.} The computing nodes harness an Intel$^\circledR$ Xeon(R) CPU E5-2620 (2.00GHz, RAM 64GB). Finally, snapshots of oil concentration $\boldsymbol{\alpha}_t$ and velocities $\bV_{x,t}, \bV_{y,t}, \bV_{z,t}$ 
in the $x,y,z$ axes respectively (i.e., $\boldsymbol{U}_{k,t} = [V_{x,t}, V_{y,t}, V_{z,t}] $) can be generated from the \ac{CFD} model to describe the two-phase flow dynamics. In this study, we are interested in building a machine learning surrogate model for predicting the evolution of $\boldsymbol{\alpha}_t$ along the test section. The training of autoencoders and LSTM is based on 1000 snapshots (i.e., every 0.01s) as described in Section~\ref{sec:results1}.

\subsection{Numerical results of latent surrogate modelling}
\label{sec:results1}
In this section, we compare different latent surrogate \rev{modelling techniques}, including both \ac{ROM} and \ac{RNN} approaches in the \ac{CFD} application described in Section~\ref{sec:CFD}.
\subsubsection{ROM reconstruction}
\begin{figure}[h!]
\centering
\includegraphics[width = 4.5in]{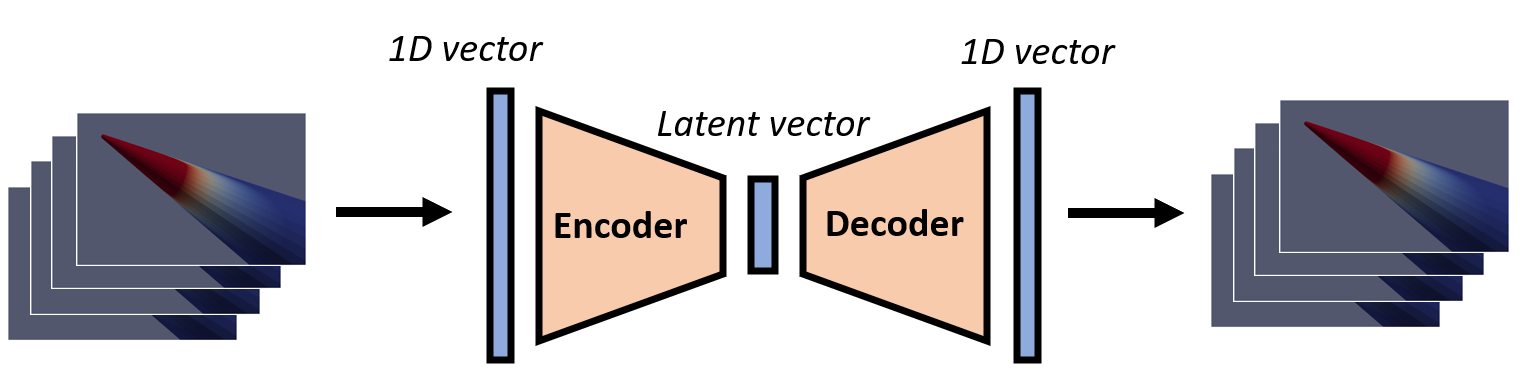} 
   \caption{\rev{Encoder-decoder modelling for the two-phase flow in the pipe.}}
   \label{fig:autoencoding}
\end{figure}

We first compare the performance of the different autoencoding approaches introduced in Section~\ref{sec:ROM}. The single-trajectory simulation data of 1000 snapshots in total are split into a training (including validation) dataset with $80\%$ of snapshots and a test dataset with the \rev{remaining} $20\%$ snapshots. Following the setup in \cite{amendola2020}, the data split is performed homogeneously where the four snapshots between \rev{two consecutive test snapshots} are used for training.  In other words, the test dataset contains the snapshots $\{ \boldsymbol{\alpha}_4, \boldsymbol{\alpha}_9, \boldsymbol{\alpha}_{14}, ..., \boldsymbol{\alpha}_{999} \}$.
Since we are dealing \rev{with cylindrical meshes} and the length of the pipe ($4m$) is much larger than its diameter ($26mm$), we decide to first flatten the snapshots to 1D vectors before auto-encoding as shown in figure~\ref{fig:autoencoding}. 

\begin{figure}[h!]
\centering
\includegraphics[width = 2.5in]{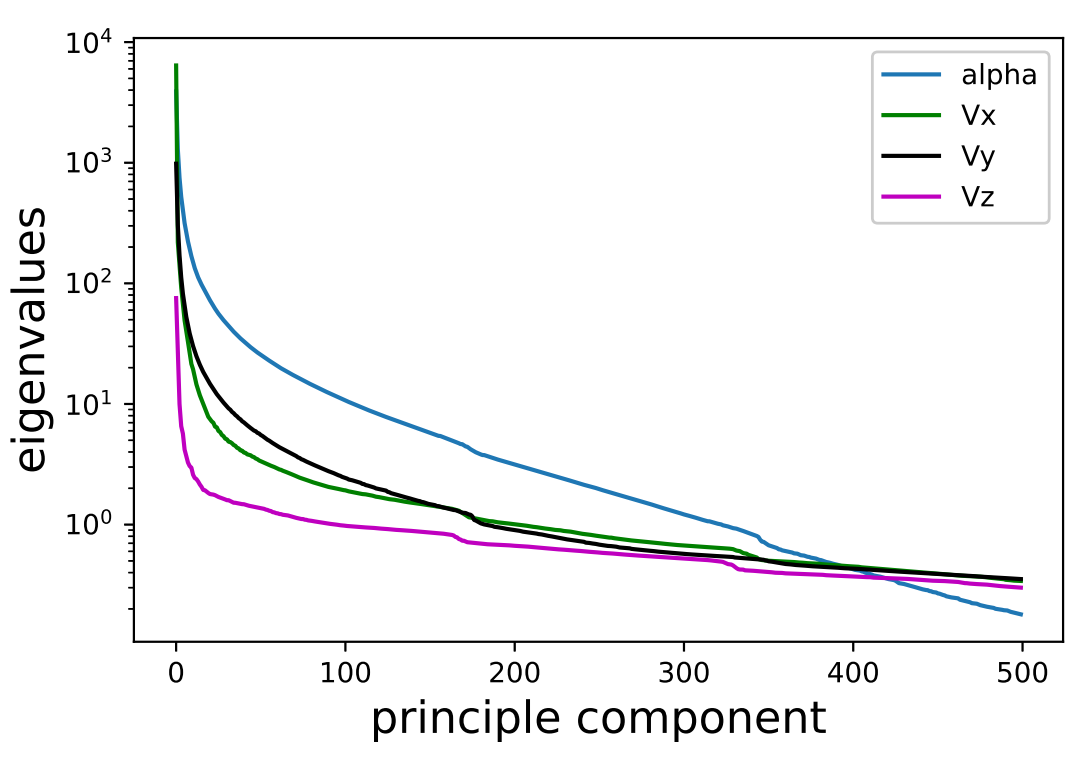} 
   \caption{Eigenvalues for $\balpha, \bV_x, \bV_y$ and $\bV_z$ on the training set, issued from one simulation.}
   \label{fig:eig}
\end{figure}

\noindent \underline{POD}\\
The distribution of the eigenvalues respectively for  $\balpha$, normalised $ \bV_x,$ normalised $\bV_y$ and normalised $ \bV_z$ is shown in figure~\ref{fig:eig} while the compression accuracy $\gamma$ and rate $\rho$, as defined in equation~\eqref{eq:POD rate}, are displayed in Table~\ref{table: compression_rate} for the truncation paramater $q=30$. In this application, \ac{POD} exhibits a high compression accuracy with an extremely low compression rate on the training data set issued from one \ac{CFD} simulation. The performance on the test dataset will be further examined in Section~\ref{sec:num comp}.

\begin{table}
\centering
\caption{Compression accuracy $\gamma$ and rate $\rho$ with truncation parameter $q = 30$ for $\balpha, \bV_x, \bV_y$ and $\bV_z$}
\begin{tabular}{ccccc} \toprule
    {\textbf{Field}} 
    & {$\balpha$} & {$\bV_x$} & {$\bV_y$}  & {$\bV_z$}\\ \midrule
    {$\gamma$}  & 99.76\% & 99.99\% &  99.81\% &  96.40\%\\
    {$\rho$}  & 1.66$\times 10^{-5}$ & 1.66$\times 10^{-5}$ & 1.66$\times 10^{-5}$ & 1.66$\times 10^{-5}$\\
    \bottomrule
\end{tabular}
\label{table: compression_rate}
\end{table}

\begin{figure}[h!]
\centering
\subfloat[Original]{\includegraphics[width = 1.8in]{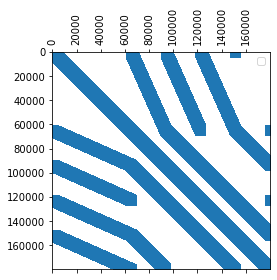}} 
\hspace{10mm}
\subfloat[Ordered]{\includegraphics[width = 1.8in]{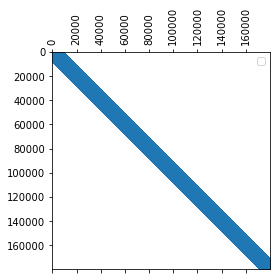}}
   \caption{\rev{Adjacency matrices before (a) and after (b) mesh reordering}}
   \label{fig:adj}
\end{figure}

\noindent\underline{1D CAE}\\
\rev{Since} the meshes have an unsquared structure and the pipe's length is much larger than the diameter, we decide to proceed with 1D CAE.
As pointed out by \cite{li2018pointcnn}, the ordering of points is crucial in CNN algorithms especially for problems with  non-square meshes. \rev{Denoting $ \mathcal{Z} = \{z_1, z_2, ..z_{n_z} \}$ the ensemble of nodes in the mesh structure, their links can be represented by the Adjacency matrix $\bA^z$ defined as}
\begin{align}
    \rev{\bA^z_{i,j} = \left\{
\begin{array}{c l}	
     & 1 \; \text{if $z_i$ is connected to $z_j$} \\
     & 0 \quad \textrm{otherwise.}
\end{array}\right.}
\end{align}
In this study, when we flatten the 3D meshes to a 1D vector, the corresponding adjacency matrix contains many non-zero values outside the diagonal band as shown in Figure~\ref{fig:adj} (a). In other words, when applying 1D CNN, \rev{the edges $\bA^z_{i,j}$ represented by the non-zero values in the adjacency matrix} can not be included in the same convolutional window thus the information of these links will be lost during the encoding. This is a common problem when dealing with unstructured or non-square meshes \cite{quilodran2021adversarial,phillips2021autoencoder}. Much effort has been devoted to finding the optimum ordering of sparse matrices for reducing the matrix band \cite{reid2006reducing, oliker2000ordering}.  In this work, we make use of the Cuthill-McKee algorithm \cite{cuthill1969reducing} based on ideas from graph theory, which is proved to be efficient for dealing with symmetric sparse matrices. The adjacency matrix \rev{for the reordered nodes} is shown in Figure~\ref{fig:adj} (b) where all non-zero elements are included in the diagonal band of width 10. We then perform the 1D CNN based on \rev{these reordered nodes}. The exact \acp{NN} structure of this 1D \ac{CAE} can be found in Table~\ref{table: CAE_structure}.

\begin{table}[h!]
\centering
\caption{NN structure of the \ac{CAE} with ordered meshes}
\begin{tabular}{ccc} \toprule
    {\textbf{Layer (type)}} 
    
    & {\textbf{Output Shape}} 
    
    & {\textbf{Activation}} \\ \midrule
     & &  \\
    {\textbf{Encoder}} & &  \\
    {Input}  
    & {$(180000, 1)$}
    & {}\\
    {\textrm{Conv 1D} $(8)$}  
    & {$(180000, 4)$}
    & {ReLu}\\
    {\textrm{Droppout} $(0.2)$}  
    & {$(180000, 4)$}
    & {}\\
    {\textrm{MaxPooling 1D} $(5)$}  
    & {$(36000, 4)$} 
    & \\
        {\textrm{Conv 1D} $(8)$}  
    & {$(36000, 4)$}
    & {ReLu}\\
    {\textrm{Droppout} $(0.2)$}  
    & {$(36000, 4)$}
    & {}\\
    {\textrm{MaxPooling 1D} $(5)$}  
    & {$(7200, 4)$} 
    & \\
        {\textrm{Conv 1D} $(8)$}  
    & {$(7200, 1)$}
    & {LeakyReLu $(0.2)$}\\
    {\textrm{AveragePooling 1D} $(5)$}  
    & {$(1440, 1)$} 
    & \\   
    {\textrm{Flatten}}  
    & {$720$} 
    & \\   
    {\textrm{Dense} $(30)$}  
    & {$30$} 
    & {ReLu}\\   
     & & \\
    {\textbf{Decoder}} & & \\
    {Input}  
    & {$30$}
    & {}\\
     {\textrm{Flatten} $(720)$}  
    & {$720$} 
    & \\  
    {\textrm{Conv 1D} $(8)$}  
    & {$(720, 1)$}
    & {ReLu}\\
    {\textrm{Upsampling} $(10)$}  
    & {$(7200, 1)$} 
    & \\
    
    {\textrm{Conv 1D} $(8)$}  
    & {$( 7200, 4)$}
    & {ReLu}\\
    {\textrm{Upsampling} $(5)$}  
    & {$(36000, 4)$} 
    & \\
    {\textrm{Conv 1D} $(8)$}  
    & {$(36000, 4)$} 
    & {LeakyReLu $(0.2)$}\\
    {\textrm{Upsampling} $(5)$}  
    & {$(180000, 1)$} 
    & {}\\
    \bottomrule
\end{tabular}
\label{table: CAE_structure}
\end{table}

%
\noindent\underline{POD AE}\\
We first apply the \ac{POD} operators to obtain the full set of PCs of $\balpha, \bV_x, \bV_y$ and $\bV_z$ respectively as described in Section~\ref{sec: POD}. Since $20\%$ of the snapshots are used for testing, we obtain 799 PCs for each variable.
Then the auto-encoding of $\balpha, \bV_x, \bV_y$ and $\bV_z$ to compressed latent variables $\tilde{\balpha}, \tilde{\bV}_x, \tilde{\bV}_y$ and $\tilde{\bV}_z$  is performed individually with the same \acp{NN} structure as displayed in Table~\ref{table: PODAE_structure}.
The training is very efficient for \ac{POD} \ac{AE} \rev{so much that it} can be easily performed on a laptop CPU in less than 15 minutes. On the other hand, 1D \ac{CAE} training takes several hours if training with the full set of snapshots. 

\begin{table}
\centering
\caption{NN structure of the \ac{POD} AE}
\begin{tabular}{ccc} \toprule
    {\textbf{Layer (type)}} 
    
    & {\textbf{Output Shape}} 
    
    & {\textbf{Activation}} \\ \midrule
    {\textbf{Encoder}} & &  \\
    {Input}  
    & {$799$}
    & {}\\
    {\textrm{Dense} $(128)$}  
    & {$128$}
    & {LeakyReLu(0.3)}\\
    {\textrm{Dense} $(30)$} 
    & {$30$}
    & {LeakyReLu(0.3)}\\
    
     & & \\
    {\textbf{Decoder}} & & \\
    {Input}  
    & {$30$}
    & {}\\
     {\textrm{Dense} $128$}  
    & {$128$}
    & {LeakyReLu(0.3)}\\
     {\textrm{Dense} $799$}  
    & {$799$}
    & {LeakyReLu(0.3)}\\
    \bottomrule
\end{tabular}
\label{table: PODAE_structure}
\end{table}

\begin{figure}[h!]
\centering
\includegraphics[width = 6.in]{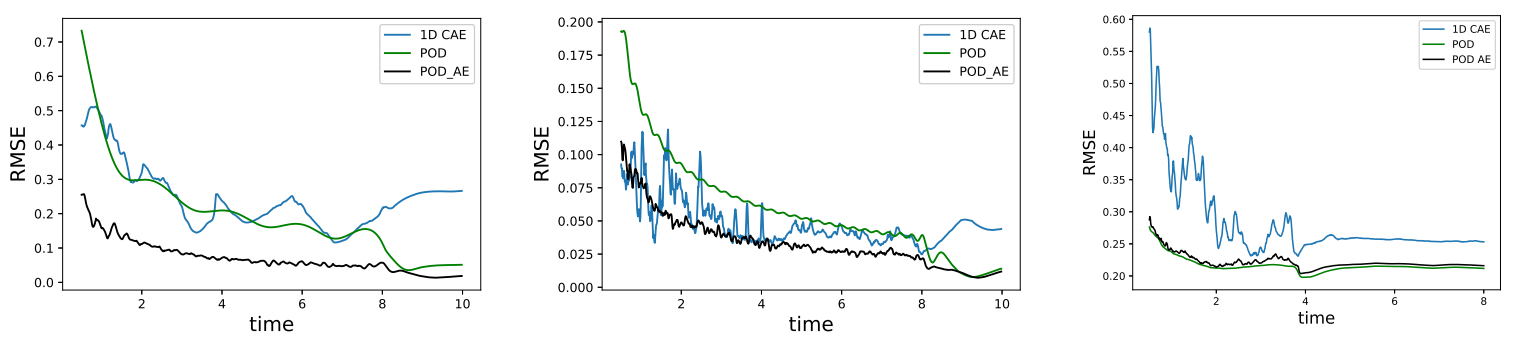}
   \caption{Comparison of reconstruction errors \rev{of oil concentration $\alpha$} using different auto-encoder approaches. \rev{Figures (a) and (b) are evaluated on the simulation data of $U_m = 0.52$ (i.e., the first row of table \ref{Table2}) while figure (c) is evaluated on the simulation data of $U_m = 0.52$ (i.e., the second row of table \ref{Table2})} }
   \label{fig:recon_error}
\end{figure}

\noindent \underline{Numerical comparison}\\
\label{sec:numerical_AE}
The relative mean square error (RMSE) \rev{for the oil concentration $\alpha$} of different \ac{ROM} reconstructions is illustrated in figure~\ref{fig:recon_error} on the \ac{CFD} simulations. The first simulation (figure~\ref{fig:recon_error}(a,b)) includes both training ($80\%$) and test ($20\%$) data while the second simulation (figure~\ref{fig:recon_error}(c)) consists of purely unseen test data. In order to further inspect the \ac{ROM} accuracy against the dimension of the latent space (i.e., the truncation parameter), we show in figure~\ref{fig:recon_error} the performance for both $q=5$ (a) and $q=30$ (b,c). It can be clearly observed that the \ac{POD} and 1D \ac{CAE} (\rev{with reordered nodes}) are out-performed by \ac{POD} \ac{AE} in terms of both average accuracy and temporal robustness for the first \ac{CFD} simulation data. For all \ac{ROM} approaches, a higher dimension of the latent space ($5 \longrightarrow 30$) can significantly enhance the reconstruction. In the case of \ac{POD} AE, the RMSE has been reduced from around $10\%$ to around $3\%$. We thus choose to use the \ac{POD} \ac{AE} approach for computing the latent surrogate model in this work. \rev{As expected, the RMSE evaluated on the second simulation dataset is larger than the first one. In figure~\ref{fig:recon_error}(c), the POD and POD AE show a better generalizability compared to the 1D CAE, which confirms our choice of POD AE in this application.}
\label{sec:num comp}

\subsubsection{LSTM surrogate model}

\begin{figure}[h!]
\centering
\includegraphics[width = 4.5in]{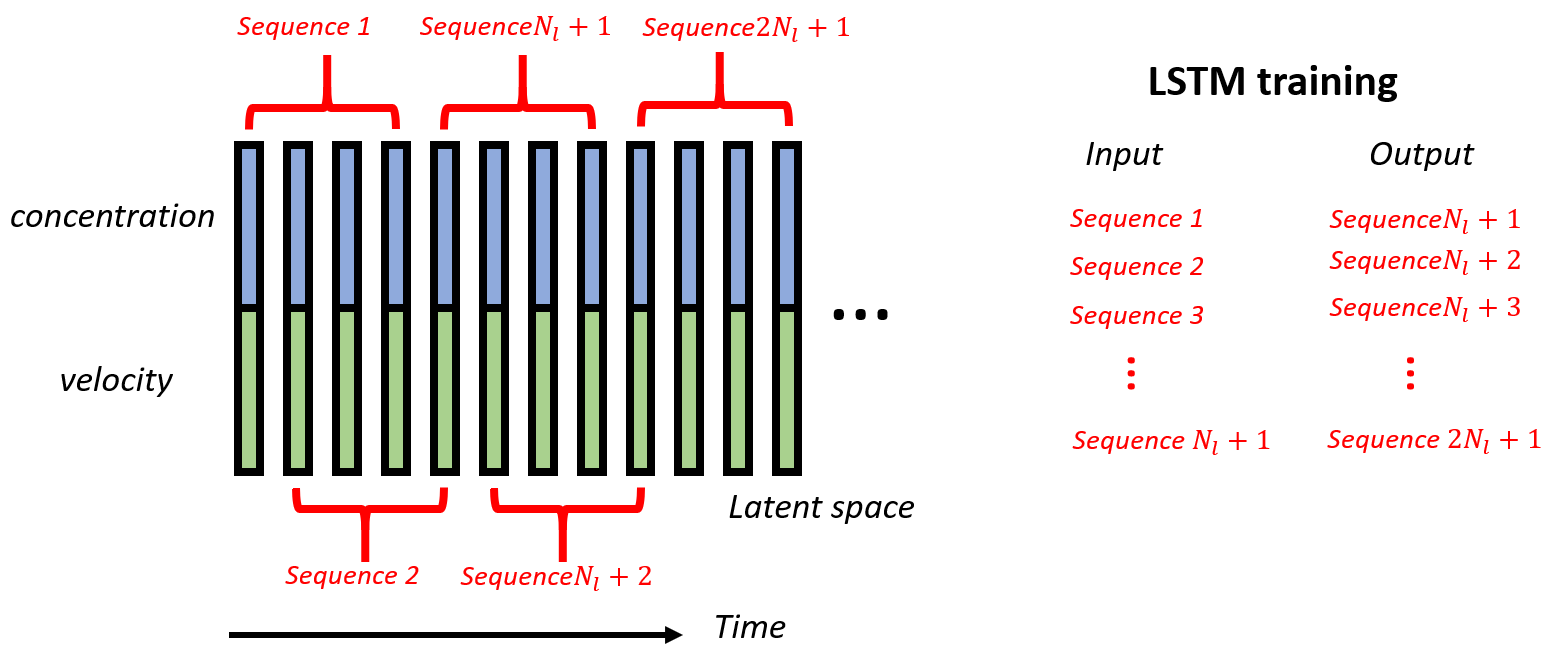} 
   \caption{LSTM training in the latent space for a joint model of concentration and velocity}
   \label{fig:LSTM_train}
\end{figure}

\begin{figure}[h!]
\centering
\includegraphics[width = 4.5in]{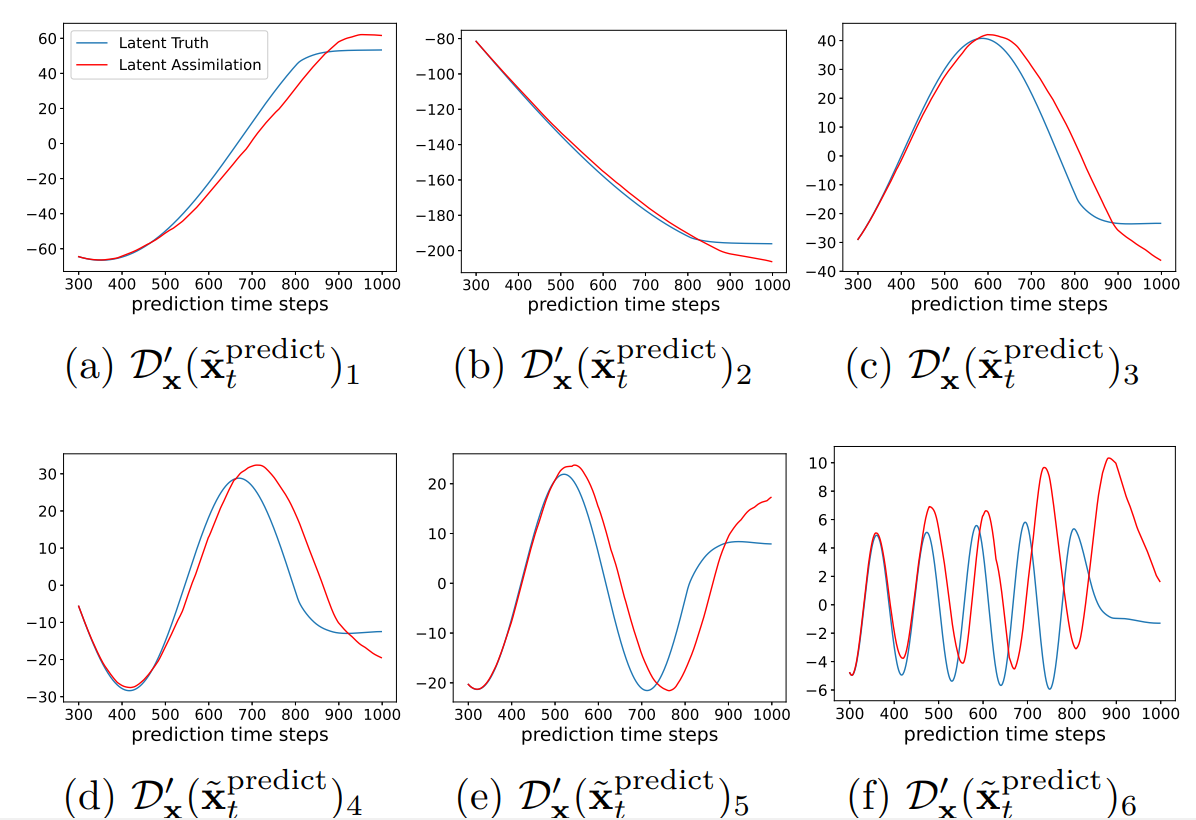}
   \caption{The LSTM prediction of reconstructed \ac{POD} coefficients (i.e., $\mathcal{D}'_{\bx}(\tilde{\bx}^{\textrm{predict}}_t)$) with joint LSTM 30to30 surrogate model}
   \label{fig:LSTM_joint}
\end{figure}

\begin{figure}[h!]
\centering
\includegraphics[width = 4.5in]{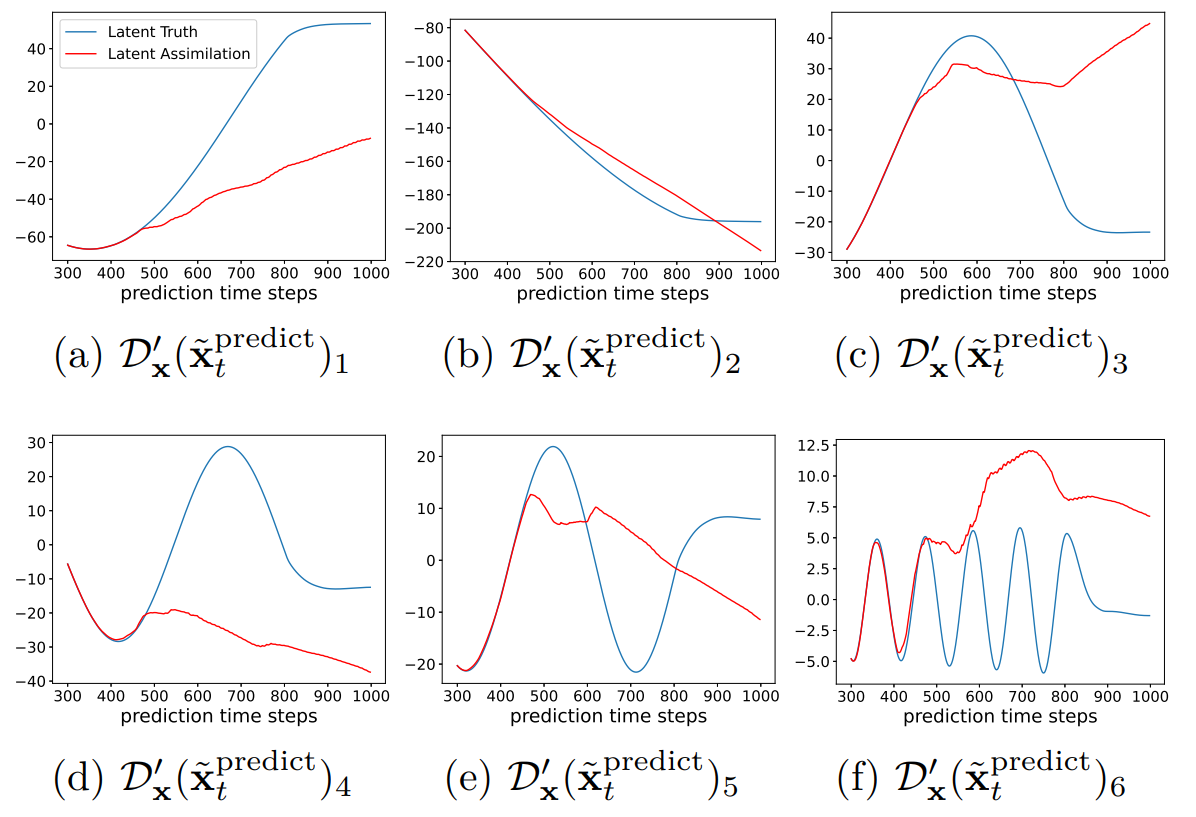}
   \caption{The LSTM prediction of reconstructed \ac{POD} coefficients (i.e., $\mathcal{D}'_{\bx}(\tilde{\bx}^{\textrm{predict}}_t)$) with single LSTM 10to10 surrogate model}
   \label{fig:LSTM_con}
\end{figure}

\begin{figure}[h!]
\centering
\subfloat[CFD]{\includegraphics[width = 1.45in]{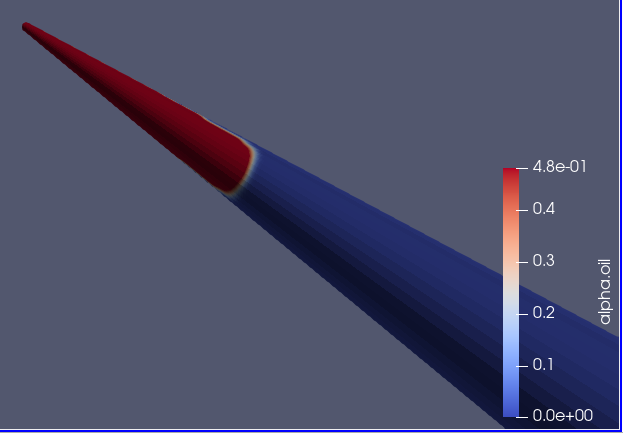}} \quad
\subfloat[LSTM joint]{\includegraphics[width = 1.45in]{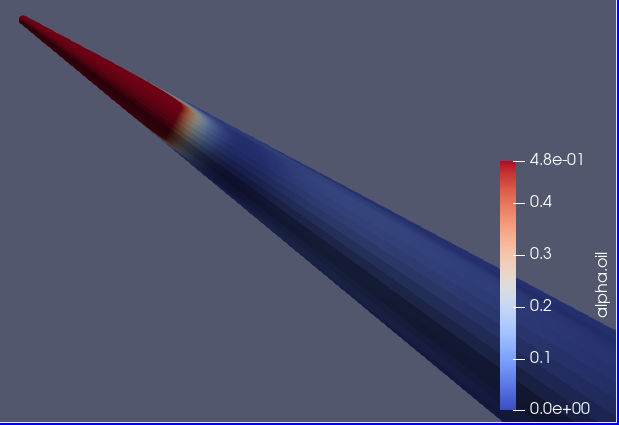}} \quad
\subfloat[LSTM single]{\includegraphics[width = 1.45in]{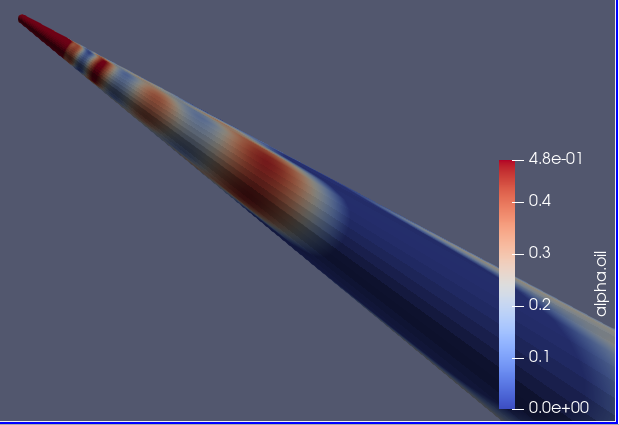}} 
   \caption{The original \ac{CFD} simulation against LSTM predictions at $t = 7s$}
   \label{fig:full_prediction}
\end{figure}

In this study, instead of classical many-to-one LSTM setting (e.g., \cite{amendola2020, nakamura2021convolutional}), we make use of a sequence-to-sequence LSTM structure to speed up the evaluation of the surrogate model. More precisely, in lieu of a single time output, the LSTM predicts a time series of latent variables with an internal connection according to the time steps. For more details about \rev{sequence-to-sequence} LSTM, interested readers are referred to the work of \cite{sutskever2014sequence}. The recent work of \cite{carta2020incremental} shows that incremental LSTM which forecasts the difference between output and input variables can significantly improve the accuracy and efficiency of \rev{the learning procedure}, especially for multiscale and multivariate systems. Therefore, we have adapted the incremental LSTM in the \rev{sequence-to-sequence} learning with
\begin{itemize}
    \item LSTM input: $\bu_\textrm{input} = [\tilde{\bx}_t, \tilde{\bx}_{t+1},..., \tilde{\bx}_{t+l_\textrm{input}-1}]$,
    \item LSTM output: $\bu_\textrm{output} = [\tilde{\bx}_{t+l_\textrm{input}}-\tilde{\bx}_{t+l_\textrm{input}-1}, \tilde{\bx}_{t+l_\textrm{input}+1}-\tilde{\bx}_{t+l_\textrm{input}},..., \tilde{\bx}_{t+l_\textrm{input}+l_\textrm{output}-1}-\tilde{\bx}_{t+l_\textrm{input}+l_\textrm{output}-2}]$,
\end{itemize}
where $l_\textrm{input}$ and $l_\textrm{output}$ denote the length of the input and the output sequences respectively. $\tilde{\bx}_{t}$ represents the latent vector encoded via the \ac{POD} \ac{AE} approach at time step $t$. The training data is generated from the simulation snapshots by shifting the beginning of the input sequence as shown in figure~\ref{fig:LSTM_train}. Similar to the setup of AEs, $80\%$ of \rev{input and output sequences} are used as training data while the remaining $20\%$ are divided into the test dataset. In this work, we implement two LSTM models where the first one includes only the encoded concentration (i.e., $\tilde{\boldsymbol{\alpha}}$) and the second one uses both concentration and velocity variables (i.e., $ \tilde{\boldsymbol{\alpha}}, \tilde{\bV}_x, \tilde{\bV}_y, \tilde{\bV}_z$) as illustrated in figure~\ref{fig:LSTM_train}. We set $l_\textrm{intput} = l_\textrm{output} = 30$ for the joint LSTM model (i.e., the one including the velocity data), meaning that 33 iterative applications of LSTM are required to predict the whole \ac{CFD} model. On the other hand, the single concentration model is trained using a LSTM 10to10 (i.e., $l_\textrm{intput} = l_\textrm{output} = 10$) since the instability of the single model doesn't support long range predictions, which will be demonstrated later in this section. The exact \acp{NN} structure of the  joint LSTM model is shown in table~\ref{table: lstm_structure} where the \rev{sequence-to-sequence} learning is performed. On the other hand, the single conceration model is implemented thanks to the \textit{RepeatVector} layer. The reconstructed \rev{principle components} via LSTM prediction (i.e., $\mathcal{D}'_{\bx}(\tilde{\bx}^{\textrm{predict}}_t)$ following the notation in Section~\ref{sec:POD AE}) against compressed ground truth (i.e., $\bL_{\bx}^T (\bx)$) are shown in figures~\ref{fig:LSTM_joint} and~\ref{fig:LSTM_con}. As observed in figure~\ref{fig:LSTM_con}, the latent prediction is accurate until around 200 time steps ($2s$) for all eigenvalues. However, a significant divergence can be observed just after $t=2s$ for most principal components due to the accumulation of prediction error. On the other hand, the joint LSTM model with similar \acp{NN} structures exhibits a much more robust prediction performance despite that some temporal gap can still be observed. The reconstructed prediction of oil concentration $\boldsymbol{\alpha}$ at $t = 7s$ (i.e. $\mathcal{D}'_{\bx}(\tilde{\bx}^{\textrm{predict}}_{t=700})$), together with the \ac{CFD} simulation of $\boldsymbol{\alpha}_{t=700}$ are illustrated in figure~\ref{fig:full_prediction}. The joint LSTM model predicts reasonably well the \ac{CFD} simulation with a slight delay of the oil dynamic while the prediction of the single LSTM model diverges at $t=7s$. These results are coherent with our analysis of figure~\ref{fig:LSTM_joint} and~\ref{fig:LSTM_con}.
In summary, although the objective here is to build a surrogate model for simulating the oil concentration, it is demonstrated numerically that more physics information can improve the prediction performance. \rev{The computational time of both LSTM surrogate models (on a Laptop CPU) and \ac{CFD} (with parallel computing mode) approaches for the entire simulation is illustrated in table~\ref{table: LSTM_time}}. For both LSTM models the online prediction takes place from t=1s ($100^\textrm{th}$ time step) until t = 10s ($1000^\textrm{th}$ time step) where the first 100 time steps of exact encoded latent variables are provided to 'warm up' the prediction system. From table~\ref{table: LSTM_time}, one observes that the \rev{online} computational time of LSTM surrogate models is around 1000 times shorter compared to the \ac{CFD}. Table~\ref{table: LSTM_time} also reveals the fact that a longer prediction sequence in \rev{sequence-to-sequence} LSTM can significantly reduce \rev{the online prediction complexity}. \rev{As shown in Table~\ref{table: LSTM_time}, the offline computation of both approaches is also very fast thanks to the training efficiency of \ac{POD} \ac{AE}.}

\begin{table}
\centering
\caption{Computational time of LSTM surrogate models and \ac{CFD} modelling}
\rev{\begin{tabular}{cccc} \toprule
    & LSTM 10to10
    & {LSTM 30to30} 
    & CFD\\ \midrule
    {Offline time}  
    & {$1426s$ }
    & {$1597s$ }
    & {}\\
    {Online time}  
    & {175s}
    & {124s}
    & {$\approx 40$ hours}\\
    \bottomrule
\end{tabular}}
\label{table: LSTM_time}
\end{table}

\begin{table}
\centering
\caption{LSTM structure in \ac{POD} \ac{AE} latent space for the single model (only concentration) and the joint model (concentration and velocity)}
\begin{tabular}{cccc} \toprule
    {\textbf{Layer (type)}} 
    & \makecell{\textbf{Output Shape} \\ single model} 
    & \makecell{\textbf{Output Shape} \\ joint model} 
    & {\textbf{Activation}} \\ \midrule
    {\textrm{Input}}  
    
    & {$(30,30)$}
    & {$(30,120)$}
    & {}\\
    {\textrm{LSTM}}  
    
    & {$50$}
    & {$200$}
    & {Sigmoid}\\
    {\textrm{RepeatVector}}  
    
    & {$(30,50)$} 
    & {$(30,200)$} 
    & \\
        {\textrm{LSTM}}  
    
    & {$(30,100)$}
    & {$(30,200)$}
    & {ReLu}\\
    
    {\textrm{Dense}}  
    
    & {$(30,200)$}
    & {$(30,200)$}
    & {ReLu}\\

    {\textrm{Time distributed}}  
    
    & {$(30,30)$}
    & {$(30,120)$}
    & {LeakyReLu}\\
    
    \bottomrule
\end{tabular}
\label{table: lstm_structure}
\end{table}

\section{Results: GLA approach}
\label{sec:results2}
In this section, we test the performance of the novel generalised latent assimilation algorithm on the \ac{CFD} test case of oil-water two-phase flow. The strength of the new approach proposed in this paper compared to existing \ac{LA} methods, is that \ac{DA} can be performed with heterogeneous latent spaces for state and observation data. In this section, we evaluate the algorithm performance using randomly generated observation function $\mathcal{H}$ in the full space.  
\subsection{Non-linear observation operators}
\label{sec:obs operator}
In order to evaluate the performance of the novel approach, we work with different synthetically generated non-linear observation vectors for \ac{LA}. Since we would like to remain as general as possible, we prefer not to set a particular form of the observation operator, which could promote some space-filling properties. For this purpose, we decide to model the observation operator 
with a random matrix $\textbf{H}$ acting as a binomial selection operator. The full-space transformation operator $\mathcal{H}$ consists of the selection operator $\textbf{H}$ and a marginal non-linear function $f_\mathcal{H}$.
Each observation will be constructed as the sum of a few true state variables randomly collected over the subdomain.
In order to do so, 
we introduce the notation for a subset sample $\left \{ \textsc{x}_t^*(i) \right \}_{i=1\ldots n_\textrm{sub}}$ randomly but homogeneously chosen (with replacement) with probability $P$ among the available data set $\left \{ \textsc{x}_t(k) \right \}_{k=1\ldots n=180000}$. The evaluations of the $f_\mathcal{H}$ function on the subsets (i.e., $f_\mathcal{H}(\textsc{x}_t^*)$) are summed up and the process is re-iterated $m \in \{10000, 30000\}$ times in order to construct the observations:
\begin{equation}
    y_t(j) = \sum_{i=1}^{n_j} f_\mathcal{H} (\textsc{x}_t^*(i)), \quad \text{for} \; j=1,\ldots,m,
\end{equation}
where the size $n_j$ (\rev{invariant with time}) of the collected sample used for each $j^{\text{th}}$ observation data point $y_t(j)$ is random and by construction follows a binomial distribution $\mathcal{B}(n,P)$.
As for the entire observation vector,
\begin{align}
    & \by_t = \begin{bmatrix}
        y_t(0) \\ 
        y_t(1)\\
        \vdots \\
        y_t(m-1) 
    \end{bmatrix} = \mathcal{H} (\bx_t) =  \bH f_\mathcal{H} (\bx_t) = \begin{bmatrix}
        \bH_{0,0}, \hdots \bH_{0,n-1} \label{eq:generating_obs} \\ 
        \vdots\\
        \bH_{m-1,0}, \hdots \bH_{m-1,n-1} \\
    \end{bmatrix} \begin{bmatrix}
         f_\mathcal{H} (x_t(0)) \\ 
        f_\mathcal{H} (x_t(1))\\
        \vdots \\
        f_\mathcal{H} (x_t(n-1))
    \end{bmatrix} \\ \notag
    & \text{with} \quad \bH_{i,j} = \left\{
\begin{array}{c l}	
     & 0 \quad \textrm{with probability} \quad 1-P\\
     & 1 \quad \textrm{with probability} \quad P
\end{array}\right.. 
\end{align}
Using randomly generated selection operator for generating observation values is commonly used for testing the performance of DA algorithms (e.g., \cite{cheng2019, farchi2021using}).
 In this work we choose a sparse representation with $P=0.1\%$. Once $\textbf{H}$ is randomly chosen, it is kept fixed for all the numerical experiments in this work.  Two marginal non-linear functions $f_\mathcal{H}$ are employed in this study:
 \begin{itemize}
     \item quadratic function: $f_\mathcal{H}(x) = x^2 $
     \item reciprocal function: $f_\mathcal{H}(x) = 1/(x + 0.5) $.
 \end{itemize}
 After the observation data is generated based on equation~\eqref{eq:generating_obs}, we apply the \ac{POD} \ac{AE} approach to build an observation latent space of dimension 30 with associated encoder $\mathcal{E}_y$ and decoder $\mathcal{D}_y$. \rev{In this application, the dimension of the observation latent space is chosen as 30 arbitrarily. In general, there is no need to keep the same dimension of the latent state space and the latent observation space. } Following equations~\eqref{eq:DExy} and~\eqref{eq:generating_obs}, the state variables $\tilde{\bx}_t$ and the observations $\tilde{\by}_t$ in \ac{LA} can be linked as:
 \begin{align}
     \tilde{\by}_t &= \tilde{\mathcal{H}} (\tilde{\bx}_t)  = \mathcal{E}_{\by} \circ \bH \circ f_\mathcal{H} \circ \mathcal{D}_{\bx} (\tilde{\bx}_t).
 \end{align}

\subsection{Numerical validation and parameter tuning}
\label{sec:numerical validation}
Local polynomial surrogate functions are then used to approximate the transformation operator $\tilde{\mathcal{H}} = \mathcal{E}_{\by} \circ \bH \circ f_\mathcal{H} \circ \mathcal{D}_{\bx}$ in Latent Assimilation. In order to investigate the \ac{PR} accuracy and perform the hyper-parameters tuning, we start by computing the local surrogate function at a fixed time step $t=3s$ with $(\tilde{\bx}_{300}, \tilde{\by}_{300})$. Two \ac{LHS} ensembles $\{ \tilde{\bx}_\textrm{train}^q \}_{q = 1.. 1000}$ and $\{ \tilde{\bx}_\textrm{test}^q \}_{q = 1.. 1000}$, each of 1000 sample vectors, are generated for training and validating local \ac{PR} respectively.  \rev{As mentioned previously in Section~\ref{sec:ahls}, the polynomial degree $d^{p}$ and the \ac{LHS} range $r_s$ are two important hyper-parameters which impacts the surrogate function accuracy. $r_s$ also determines the expectation of the range of prediction errors in the \ac{GLA} algorithm.} For hyper-parameters tuning, we evaluate the root-mean-square-error (RMSE) (of $\{ \tilde{\bx}_\textrm{test}^q \}_{q = 1.. 1000}$) and the computational time of local \ac{PR} with a range of different parameters, i.e.,
\begin{align}
  &\{ \tilde{\bx}_\textrm{train}^q \}_{q = 1.. 1000} / \{ \tilde{\bx}_\textrm{test}^q \}_{q = 1.. 1000} = \textrm{LHS Sampling}_{\{d^{p},r_s,1000\}} (\tilde{\bx}_{t=300}) \notag \\
  &\textrm{for} \quad   d^{p} \in \{1,...,5 \} \quad \textrm{and} \quad r_s \in \{10\%, ..., 90\% \}.
\end{align}

\begin{figure}[h!]
\centering
\subfloat[log(RMSE)]{\includegraphics[width = 2.4in]{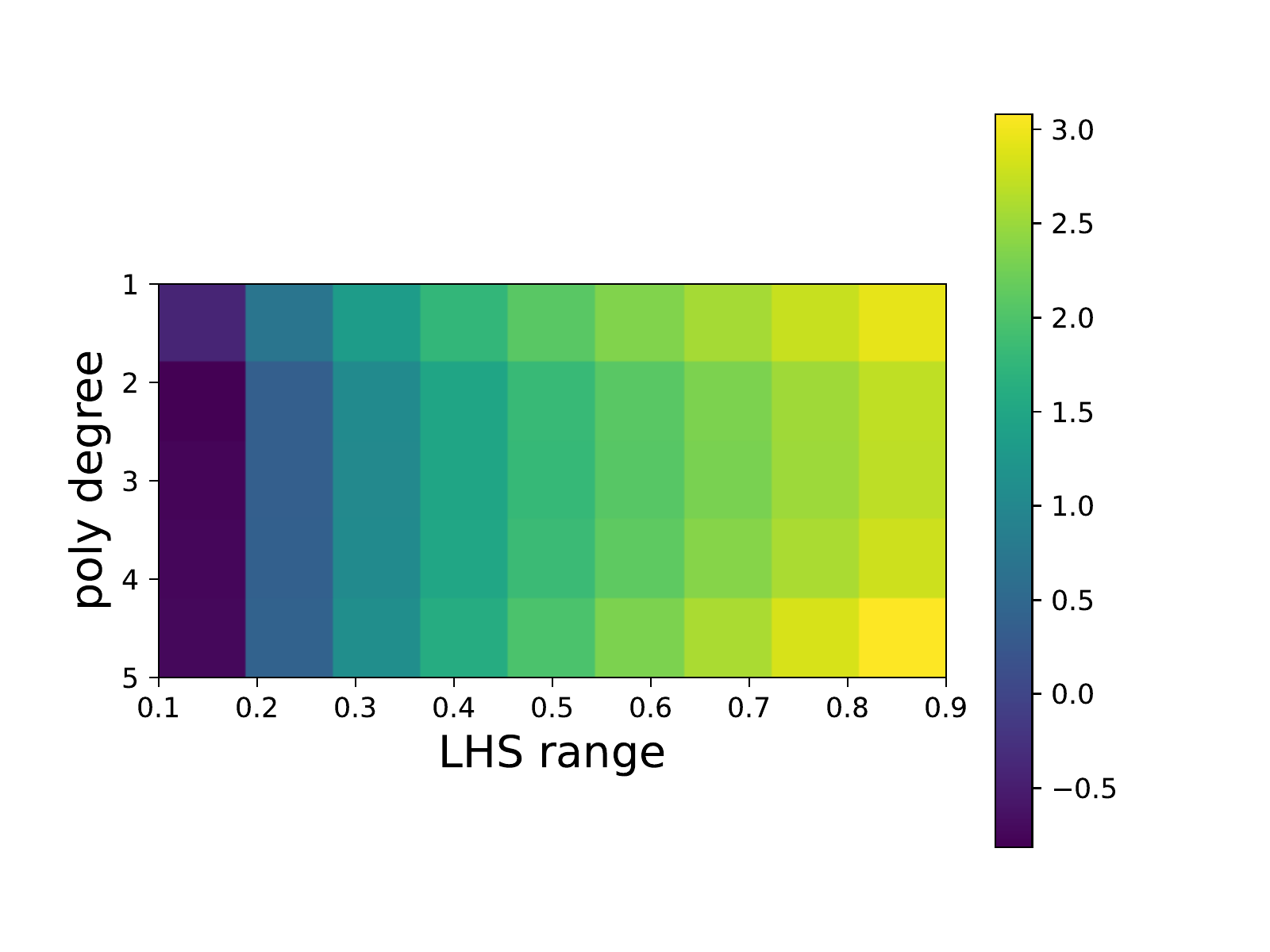}}
\subfloat[log(training time(s))]{\includegraphics[width = 2.4in]{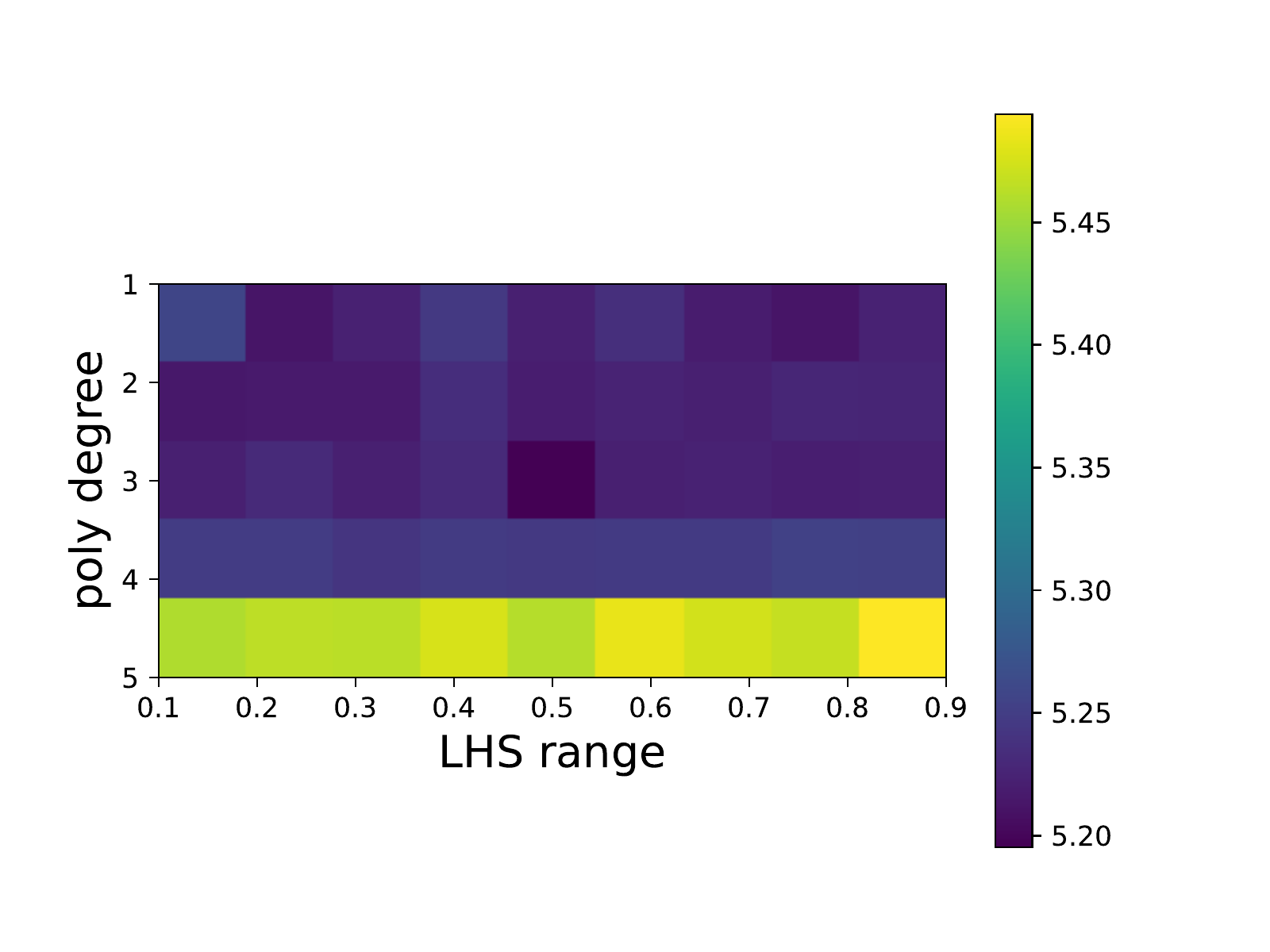}} 
   \caption{Logarithm of RMSE in the test dataset (evaluated on 1000 points) and the training time in seconds}
   \label{fig:hyperparam}
\end{figure}
\begin{figure}[h!]
\centering
\includegraphics[width = 5.in]{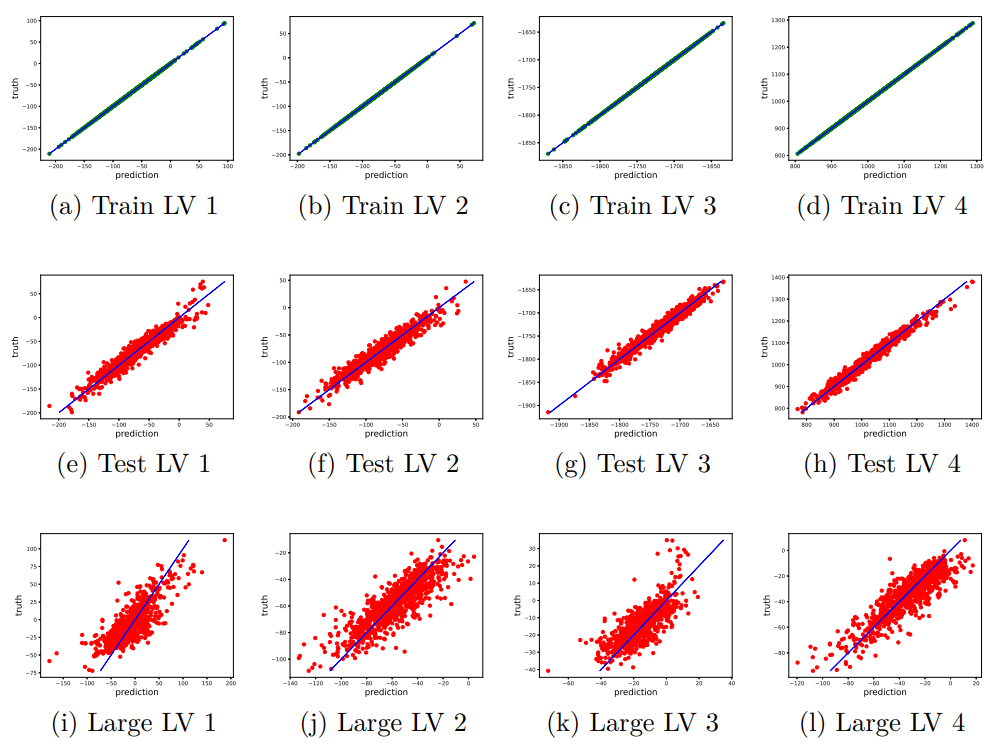}
   \caption{Latent variable prediction results in the training (a-d) and test (e-l) datasets against the true values with the polynomial degree $d^p = 4$. The \ac{LHS} sampling range is $r_s = 30\%$ for a-h and $r_s = 60\%$ for i-l.}
   \label{fig:scatter}
\end{figure}

The results are presented in figure~\ref{fig:hyperparam} with a logarithmic scale for both RMSE and computational time (in seconds). \rev{Here the quadratic function is chosen as the transformation operator to perform the tests.} Figure~\ref{fig:hyperparam}(a) reveals that there is a steady rise of RMSE against \ac{LHS} range$r_s$ . This fact shows the difficulties of \ac{PR} predictions when the input vector is far from the \ac{LHS} center (i.e., $\tilde{\bx}_{300}$) due to the high non-linearity of \acp{NN} functions. The \ac{PR}  performance for $d^p = 2,3,4$ on the test dataset $\{ \tilde{\bx}_\textrm{test}^q \}_{q = 1.. 1000}$ is more robust compared to linear predictions (i.e., $d^p = 1$), especially when the \ac{LHS} range grows. However, a phenomenon of overfitting can be noticed when $d^p \geq 5$ where an increase of prediction error is noticed. One has to make a tradeoff between prediction accuracy and application range when choosing the value of $r_s$. In general, \ac{PR} presents a good performance with a relative low RMSE (with an upper bound of $e^{3} = 20.08$) \rev{given that} $\vert \vert\tilde{\bx}_{t=300}\vert \vert_2 = 113.07$. As for the computational time of a local \ac{PR}, it stays in the same order of magnitude for different set of parameters (from $e^{5.2} \approx 181s$ to $e^{5.5} \approx 244s$) where the cases of $d^p = 1,2,3,4$ are extremely close. Considering the numerical results shown in figure~\ref{fig:hyperparam} and further experiments in Latent Assimilation, we fix the parameters as $d^p = 4$ and $r_s = 0.3$ in this application. The \ac{PR} prediction results against the compressed truth in the latent space are shown in figure~\ref{fig:scatter} for \rev{4 different latent observations}. What can be clearly seen is that the local \ac{PR} can fit very well the $\tilde{\mathcal{H}}$ function in the training dataset (figure~\ref{fig:scatter}(a-d)) while also provides a good prediction of unseen data (figure~\ref{fig:scatter}(e-h)), which is consistent with our conclusion in figure~\ref{fig:hyperparam}. \rev{When the sampling range increases in the test dataset (figure~\ref{fig:scatter}(i-l)), it is clear that the prediction start to perform less well. This represents the case where we have under-estimated the prediction error by 100\% (i.e., $r_s = 30\%$ for training and $r_s = 60\%$ for testing). The required number of samples (i.e., $n_s=1000$) is obtained by offline experiments performed at $(\bx_{300},\by_{300})$. For different polynomial degrees $d_p \in \{ 1,2,3,4,5\}$, no significant improvement in terms of prediction accuracy on the test dataset can be observed  when the number of samples $n_s>1000$.} We have also performed other experiments at different time steps (other than $t=3s$) and obtained similar results qualitatively. 

\subsection{Generalised Latent Assimilation}
In this section, we illustrate the experimental results of performing variational Generalised \ac{LA} with the \ac{POD} \ac{AE} reduced-order-modelling and the LSTM surrogate model. The loss functions in the variational methods are computed thanks to the local polynomial surrogate functions. The obtained results are compared with \ac{CFD} simulations both in the low dimensional basis and the full physical space.

\subsubsection{GLA with a quadratic operator function }
\label{sec:LA_2}
Following the setup in Section~\ref{sec:obs operator}, the full-space observation operator is computed with a binomial random selection matrix $\bH$ and quadratic marginal equation $f_\mathcal{H}(x) = x^2 $ as shown in equation~\eqref{eq:generating_obs}. Separate \ac{POD} \acp{AE} (i.e., $\mathcal{E}_{\by}$ and $\mathcal{D}_{\by}$) are trained for encoding the observation data. The prediction of the LSTM surrogate model start at $t=3s$, i.e., the $300^\textrm{th}$ time step. Since the prediction of the joint model is made using a 30 to 30 LSTM, the \ac{LA} takes place every 1.5$s$ starting from 5.7$s$ for 30 consecutive time steps each time. In other words, the \ac{LA} takes place at time steps 570 to 599, 720 to 749 and 870 to 899, resulting in 90 steps of assimilations among 700 prediction steps.  As for the 10to10 single concentration LSTM model, since the prediction accuracy is relatively mediocre as shown in figure~\ref{fig:LSTM_con}, more assimilation steps are required. In this case the \ac{LA} takes place every 0.6$s$ starting from 5$s$ for 10 consecutive time steps each time, leading to 180 in total. For the minimization of the cost function in the variational \ac{LA} (equation~\eqref{eq_latent3dvar}), Algorithm~\ref{algo:2} is performed with the maximum number of iterations $k_{max} = 50$ and the tolerance $\epsilon = 0.05$ in each assimilation window. To increase the importance of observation data, the error covariance matrices in Algorithm~\ref{algo:1} are fixed as:
\begin{align}
    \tilde{\bB} = \bI_{30} \quad \textrm{and} \quad \tilde{\bR} = 0.1 \times \bI_{30},
\end{align}
where $\bI_{30}$ denotes the identity matrix of dimension 30.\\

\begin{figure}[h!]
\centering
\includegraphics[width = 4.5in]{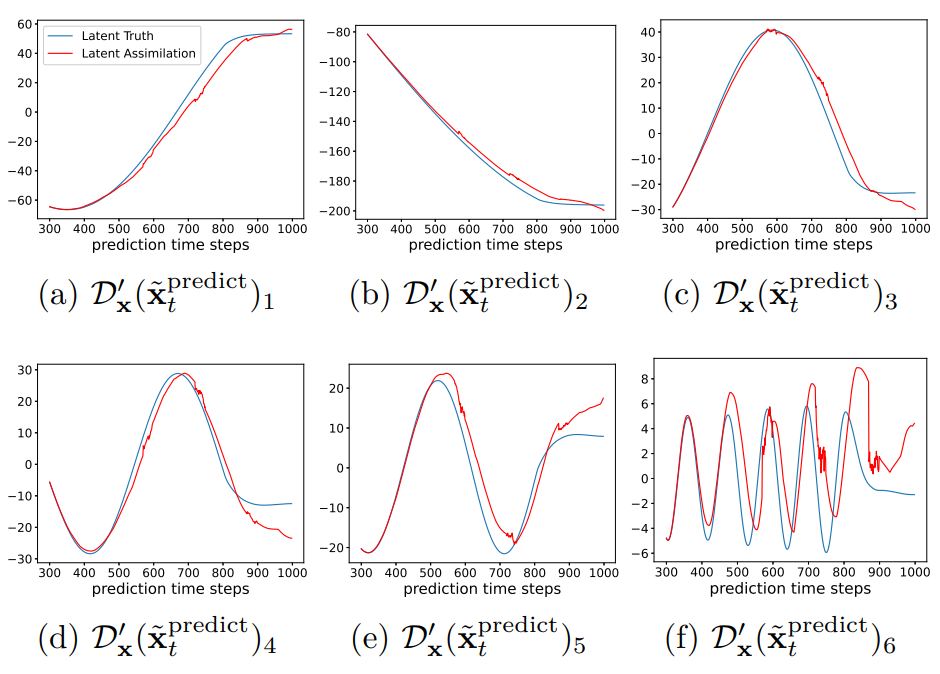}
   \caption{The \ac{LA} of reconstructed \rev{\ac{POD} coefficients} (i.e., $\mathcal{D}'_{\bx}(\tilde{\bx}^{\textrm{predict}}_t)$) with joint LSTM 30to30 surrogate model and quadratic observation function. \rev{Results of the same experiment without \ac{GLA} is shown in figure \ref{fig:LSTM_joint}}}
   \label{fig:LA_2_30}
\end{figure}

\begin{figure}[h!]
\centering
\includegraphics[width = 4.5in]{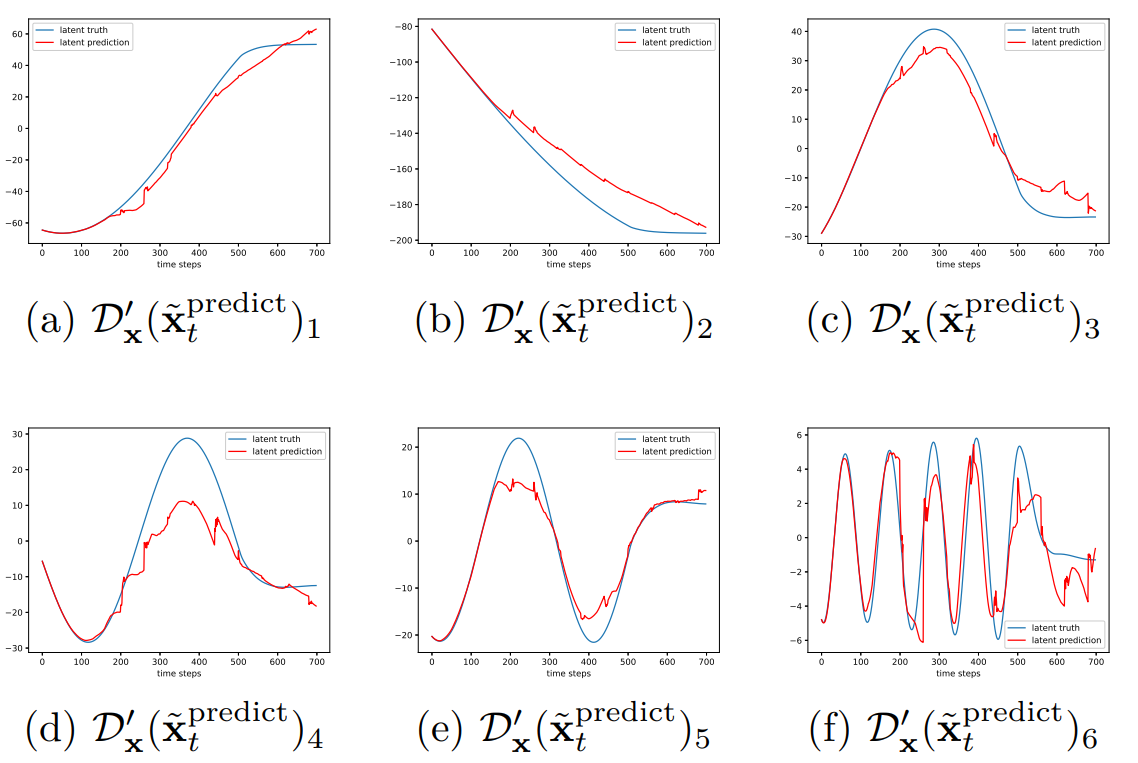}
   \caption{The \ac{LA} of reconstructed \rev{\ac{POD} coefficients} (i.e., $\mathcal{D}'_{\bx}(\tilde{\bx}^{\textrm{predict}}_t)$) with single LSTM 10to10 surrogate model and quadratic observation function. \rev{Results of the same experiment without \ac{GLA} is shown in figure \ref{fig:LSTM_con}}}
   \label{fig:LA_2_10}
\end{figure}

The Latent assimilation of \rev{reconstructed principle components} (i.e., $\mathcal{D}'_{\bx}(\tilde{\bx}^{\textrm{predict}}_t)$) against the compressed ground truth is illustrated in figure~\ref{fig:LA_2_30} and~\ref{fig:LA_2_10} for the joint and single LSTM surrogate model respectively. The red curves include both prediction and assimilation results starting at $t = 3s$ (i.e., $300^\textrm{th}$ time step). What can be clearly observed is that, compared to pure LSTM predictions shown in figure~\ref{fig:LSTM_joint} and~\ref{fig:LSTM_con}, the mismatch between predicted curves and the ground truth (CFD simulation) can be considerably reduced by the novel generalised \ac{LA} technique, especially for the single LSTM model. As for the joint LSTM surrogate model (figure~\ref{fig:LA_2_30}), the improvement is significant for $\mathcal{D}'_{\bx}(\tilde{\bx}^{\textrm{predict}}_t)_4, \mathcal{D}'_{\bx}(\tilde{\bx}^{\textrm{predict}}_t)_5, \textrm{and } \mathcal{D}'_{\bx}(\tilde{\bx}^{\textrm{predict}}_t)_6$.  
These results show that the novel approach can well incorporate real-time observation data with partial and non-linear transformation operators that the state-of-the-art \ac{LA} can not handle. Prediction/assimilation mismatch in the full physical space will be discussed later in Section~\ref{sec:error_full_space}.

\subsubsection{GLA  with a reciprocal operator function }
\label{sec:LA_invreg}
\begin{figure}[h!]
\centering
\includegraphics[width = 4.5in]{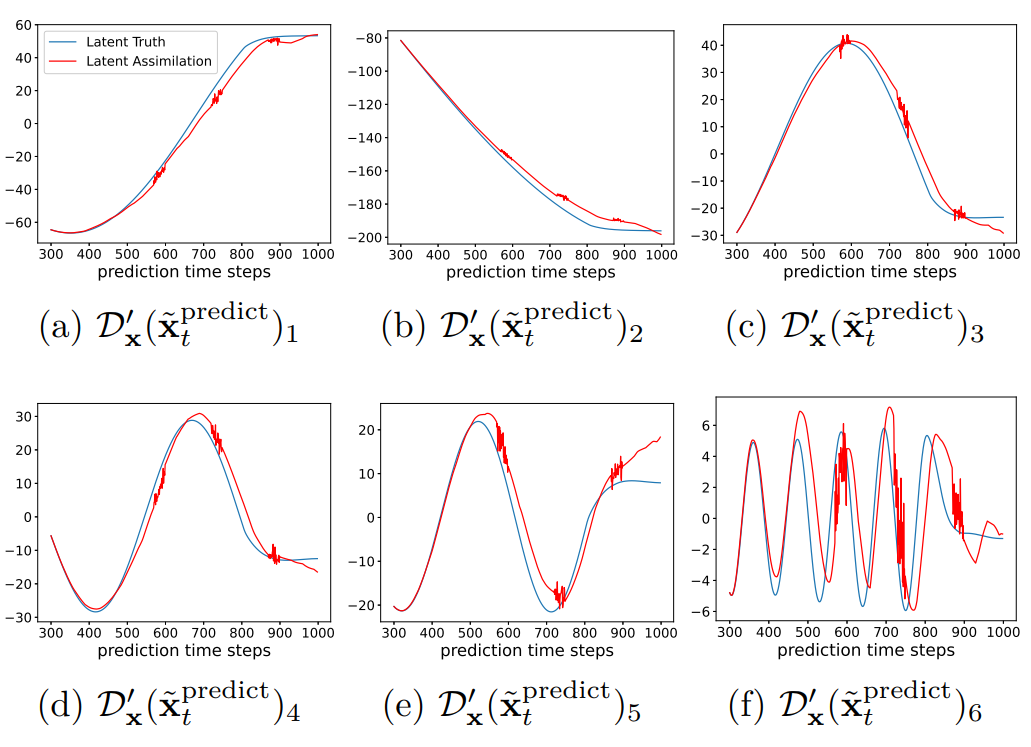} 
   \caption{The \ac{LA} of reconstructed \rev{\ac{POD} coefficients} (i.e., $\mathcal{D}'_{\bx}(\tilde{\bx}^{\textrm{predict}}_t)$) with joint LSTM 30to30 surrogate model and reciprocal observation function}
   \label{fig:LA_inv_30}
\end{figure}

\begin{figure}[h!]
\centering
\includegraphics[width = 4.5in]{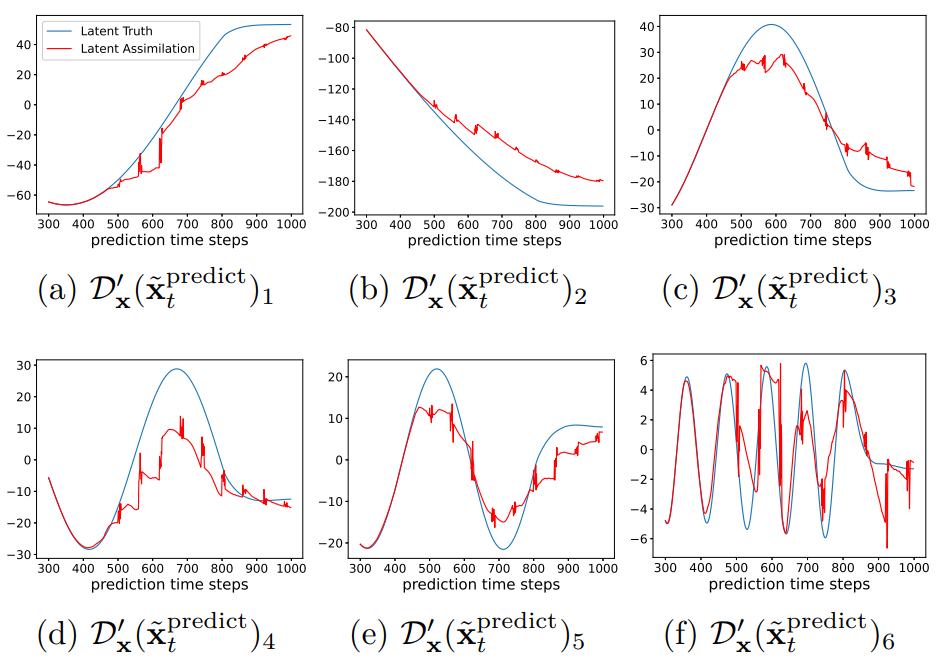}
   \caption{The \ac{LA} of reconstructed \rev{\ac{POD} coefficients} (i.e., $\mathcal{D}'_{\bx}(\tilde{\bx}^{\textrm{predict}}_t)$) with single LSTM 10to10 surrogate model and reciprocal observation function}
   \label{fig:LA_inv_10}
\end{figure}

Here we keep the same assimilation setup as in Section~\ref{sec:LA_2} in terms of assimilation accuracy and error covariances specification. Instead of a quadratic function, the observation data are generated using the reciprocal function $f_\mathcal{H}(x) = 1/(x + 0.5) $ in the full space as described in Section~\ref{sec:obs operator}. Therefore, new autoencoders are trained to compress the observation data for $\boldsymbol{\alpha}_t, \bV_{x,t}, \bV_{y,t}, \bV_{z,t}$ to latent spaces of dimension 30. The results of predicted/assimilated \rev{\ac{POD} coefficients} $\mathcal{D}'_{\bx}(\tilde{\bx}^{\textrm{predict}}_t)$ are shown in figure~\ref{fig:LA_inv_30} and~\ref{fig:LA_inv_10}. Similar conclusion \rev{can be drawn} as in Section~\ref{sec:LA_2}, that is, the generalised \ac{LA} approach manages to correctly update the LSTM predictions (for both joint and single models) on a consistent basis. \rev{Some non-physical oscillatory behaviours can be observed in figure~\ref{fig:LA_2_30}-\ref{fig:LA_inv_10}. This is due to the application of \ac{LA} which modified the dynamics in the latent space.} Comparing the assimilated curves using quadratic and reciprocal observation functions, the latter is slightly more chaotic due to the fact that reciprocal functions, when combined with DL encoder-decoders (as shown in figure~\ref{fig:PR}) can be more difficult to learn for local polynomial surrogate functions.

\subsubsection{Prediction error in the latent and the full space}
\label{sec:error_full_space}

\begin{figure}[h!]
\centering
\includegraphics[width = 4.5in]{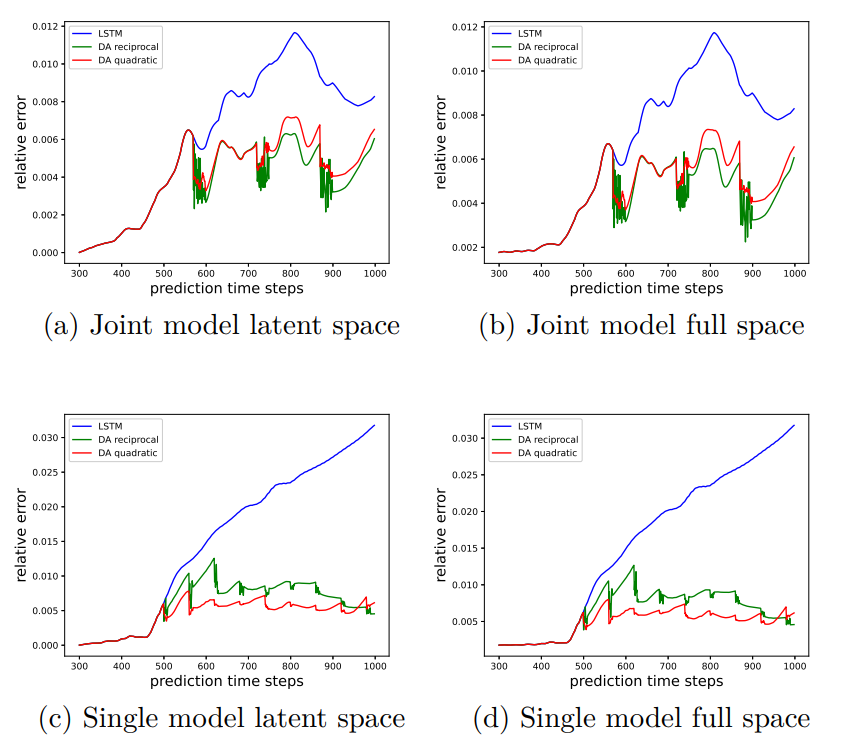}
   \caption{Prediction relative error in the latent and the full space.}
   \label{fig:relative_error}
\end{figure}

\begin{figure}[h!]
\centering
\subfloat[single LSTM (quadratic)]{\includegraphics[width = 1.45in]{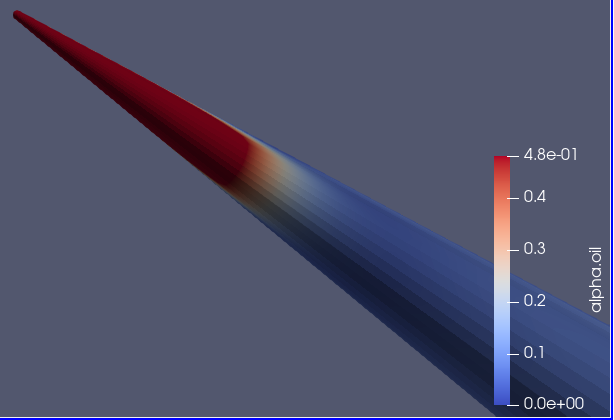}} \quad
\subfloat[single LSTM (reciprocal)]{\includegraphics[width = 1.45in]{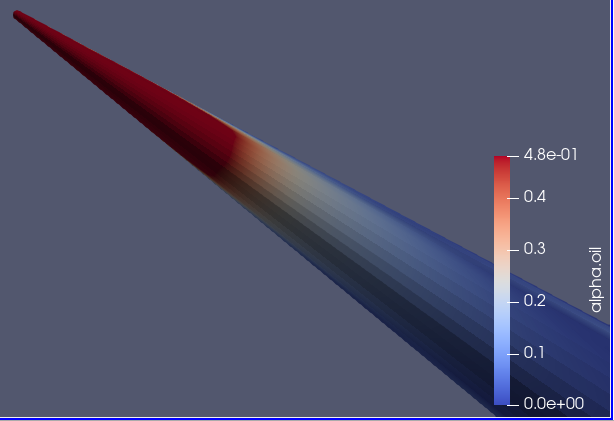}}  \\
\subfloat[joint LSTM (quadratic)]{\includegraphics[width = 1.45in]{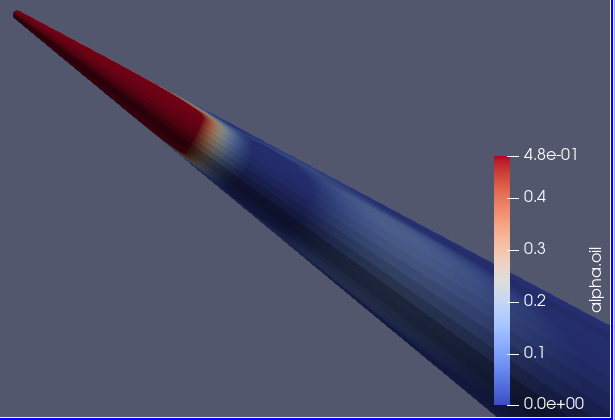}} \quad
\subfloat[joint LSTM (reciprocal)]{\includegraphics[width = 1.45in]{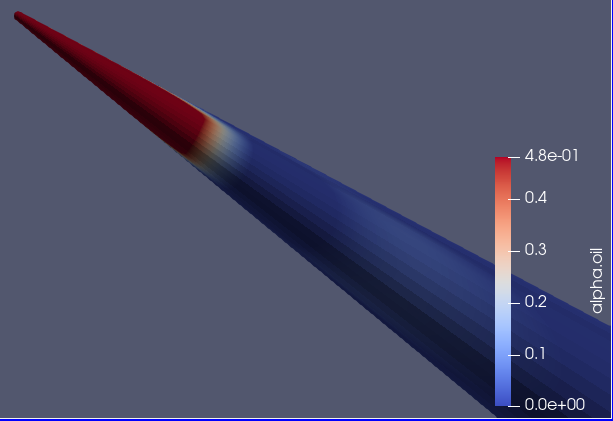}} 
   \caption{Prediction in the full \ac{CFD} space after \ac{LA} with quadratic (a,c) and reciprocal (b,d) observation functions }
   \label{fig:full_correction}
\end{figure}

In this section, we illustrate the evolution of the global prediction/assimilation errors and the forecasting of the global physical field based on the results obtained in Section~\ref{sec:LA_2} and~\ref{sec:LA_invreg}. The relative $L_2$ error in the latent space and the full space of the concentration, i.e., 
\begin{align}
    \frac{\vert \vert\bL^T_{\bx} \boldsymbol{\alpha}_t - \mathcal{D}'_{\bx}(\tilde{\bx}^{\textrm{predict}}_t)\vert \vert_2}{\vert \vert\bL^T_{\bx} \boldsymbol{\alpha}_t \vert \vert_2} \quad \textrm{and} \quad     \frac{\vert \vert\boldsymbol{\alpha}_t - \bL_{\bx} \mathcal{D}'_{\bx}(\tilde{\bx}^{\textrm{predict}}_t)\vert \vert_2}{\vert \vert\boldsymbol{\alpha}_t\vert \vert_2},
\end{align}
for both joint and single models is shown in figure~\ref{fig:relative_error}. The evolution of the relative error in the global space is consistent with our analysis in figure~\ref{fig:LA_2_30}-\ref{fig:LA_inv_10} for decoded \rev{\ac{POD} coefficients}. The \ac{LA} with quadratic (in red) and reciprocal (in green) observation operators can significantly reduce \rev{the relative error} as compared to the original LSTM model (in blue).  More importantly, the DA does not only impact the estimation of current time steps, it improves also future predictions after assimilation, thus demonstrating the stability of the proposed approach. The prediction error in the latent space and the full physical space share very similar shapes for both single and joint models, showing that the \ac{ROM} reconstruction errors are dominated by the LSTM prediction error. The reconstructed model prediction/assimilation in the full space at $t=7s$ is shown in figure~\ref{fig:full_correction}. Compared to figure~\ref{fig:full_prediction}, the prediction of the single LSTM model (figure~\ref{fig:full_correction} (a-b)) can be greatly improved with an output much more realistic and closer to the \ac{CFD} simulation (figure~\ref{fig:full_prediction} (a)). As for the joint model, the initial delay 
of the oil dynamic can also be well corrected thanks to the variational \ac{LA} approach despite some noises can still be observed. In summary, the novel \ac{LA} technique with local polynomial surrogate function manages to improve the current assimilation reconstruction, and more importantly future predictions of latent LSTM. \rev{ The optimization of equation~\eqref{eq_latent3dvar} is implemented using the ADAO~\cite{adaotrue} package where the maximum number of iterations and the stopping tolerance of the BFGS algorithm are fixed as 50 and $0.01$, respectively. }

\section{Conclusion and future work}
\label{sec:conclusions}
Performing DA with simulation and observation data encoded in heterogeneous latent spaces is an important challenge since background and observation quantities are often different in real DA scenarios.
On the other hand, it is extremely difficult, if not infeasible, to apply directly classical variational DA approaches due to the complexity and non-smoothness of the \acp{NN} function which links different latent spaces. In this paper, we introduce a novel algorithm, named generalised Latent Assimilation, which makes use of a polynomial surrogate function to approximate the \acp{NN} transformation operator in a neighbourhood of the background state. Variational DA can then be performed by computing an observation loss using this local polynomial function. This new method promotes a much more flexible use of \ac{LA} with machine learning surrogate models. A theoretical analysis is also given in the present study, where an upper bound of the approximation error of the DA cost function (evaluated on the true state) is specified. \rev{Future work can further focus on the minimization error related to the surrogate loss function in \ac{GLA}.} The numerical tests in the high-dimensional \ac{CFD} application show that the proposed approach can ensure both the efficiency of the \acp{ROM} and the accuracy of the assimilation/prediction. \rev{In this study, the training and the validation for both \ac{ROM} and LSTM are performed using a single \ac{CFD} simulation with well separated training and testing datasets. 
Future work will investigate to build robust models for both autoencding and machine learning prediction using multiple \ac{CFD} simulations as training data. However, building such training dataset can be time-consuming due to the complexity of the \ac{CFD} code.} The local polynomial surroagate function is computed relying on LHS samplings in this work. Other sampling strategies, such as Gaussian perturbations, can also be considered. \rev{Representing model or observation error (originally in the full space) in the latent space is challenging due to the non-linearity of \acp{ROM}. Future work can also be considered to enhance the error covariance specification in the latent space by investigating, for instance, uncertainty propagation from the full physical space to the latent space, posterior error covariance tuning (e.g., \cite{Desroziers2005, cheng2019, cheng2021observationRNN}) or Ensemble-type \cite{Evensen1994} DA approaches.  }

\clearpage
\section*{Main Notations}
\begin{table*}[ht!]
    \centering
    \begin{tabular}{ p{3.5cm} p{15cm}}
$\bx_t$ & state vector in the full space at time $t$\\
$\tilde{\bx}_t$ & encoded state in the latent space at time $t$\\
$\tilde{\bx}_{b,t}$ & background (predicted) state in the latent space at time $t$\\
$\tilde{\bx}_{a,t}$ & analysis (assimilated) state in the latent space at time $t$\\
${\bx}_{\textrm{true},t}/\tilde{\bx}_{\textrm{true},t}$ & true state vector in the full/latent space at time $t$\\
${\bx}^r_{\textrm{POD}},{\bx}^r_{\textrm{CAE}},{\bx}^r_{\textrm{POD AE}}$ & reconstructed state in the full space\\
$\by_t$ & observation vector in the full space at time $t$ \\
$\tilde{\by}_t$ & encoded observation vector in the latent space at time $t$\\

$\mathcal{E}_x,\mathcal{E}_y$ & encoder for state/observation vectors\\
$\mathcal{D}_x,\mathcal{D}_y$ & decoder for state/observation vectors\\
$\tilde{\bL}_{\bX,q}$ & POD projection operator with truncation parameter $q$\\
$\mathcal{H}_t$ & transformation operator in the full physical space\\
$\tilde{\mathcal{H}}_t$ & transformation operator linking different latent spaces\\
$\tilde{\mathcal{H}}^p_t$ & approximated transformation operator in GLA\\
$\tilde{\bB}_{t}, \tilde{\bR}_{t}$ & error covariance matrices in the latent space\\
    \end{tabular}
\end{table*}

\section*{Contribution statement}
S.Cheng and R.Arcucci conceived the presented idea. S.Cheng developed the theory and performed the ML and DA computations. J.Chen implemented the CFD computation and C.Anastasiou performed the physics experiments for initial conditions. O.Matar, R.Arcucci, Y-K.Guo, P.Angeli and C.Pain supervised the findings of this work. S.Cheng took the lead in writing the manuscript. All authors discussed the results and contributed to the final manuscript.

\section*{Acknowledgements}
 This research is funded by the EP/T000414/1  PREdictive Modelling with QuantIfication of UncERtainty for MultiphasE Systems (PREMIERE). This work is partially supported by the Leverhulme Centre for Wildfires, Environment and Society through the Leverhulme Trust, grant number RC-2018-023. This work is partially supported by the CAS scholarship. This work is partially supported by RELIANT (EP/V036777/1), INHALE (EP/T003189/1), Wave-Suite (EP/V040235/1) and MUFFINS (EP/P033180/1).

\footnotesize{
\bibliography{main}
}
\bibliographystyle{unsrt}


\clearpage

\section*{Acronyms}
\begin{acronym}[AAAAA]
\acro{NN}{neural network}
\acro{ML}{machine learning}
\acro{LA}{Latent Assimilation}
\acro{DA}{data assimilation}
\acro{PR}{polynomial regression}
\acro{AE}{Autoencoder}
\acro{VAE}{Variational autoencoder}
\acro{CAE}{Convolutional autoencoder}
\acro{BLUE}{Best Linear Unbiased Estimator}
\acro{3D-Var}{3D Variational}
\acro{RNN}{recurrent neural network}
\acro{CNN}{convolutional neural network}
\acro{LSTM}{long short-term memory}
\acro{POD}{proper orthogonal decomposition}
\acro{PCA}{principal component analysis}
\acro{PC}{principal component}
\acro{SVD}{singular value decomposition}
\acro{ROM}{reduced-order modelling}
\acro{CFD}{computational fluid dynamics}
\acro{1D}{one-dimensional}
\acro{2D}{two-dimensional}
\acro{NWP}{numerical weather prediction}
\acro{MSE}{mean square error}
\acro{S2S}{sequence-to-sequence}
\acro{R-RMSE}{relative root mean square error}
\acro{BFGS}{Broyden–Fletcher–Goldfarb–Shanno}
\acro{LHS}{Latin Hypercube Sampling}
\acro{AI}{artificial intelligence}
\acro{DL}{deep learning}
\acro{GLA}{Generalised Latent Assimilation}
\acro{PIV}{Particle Image Velocimetry}
\acro{LIF}{Laser Induced Fluorescence}
\end{acronym}

\end{document}